\documentclass{article}
\usepackage{widetext}
\usepackage{algorithm}
\usepackage{algpseudocode}
\usepackage{PRIMEarxiv}
\usepackage{amsmath}
\usepackage[utf8]{inputenc} 
\usepackage[T1]{fontenc}    
\usepackage{hyperref}       
\usepackage{url}            
\usepackage{booktabs}       
\usepackage{amsfonts}       
\usepackage{nicefrac}       
\usepackage{microtype}      
\usepackage{lipsum}
\usepackage{fancyhdr}       
\usepackage{graphicx}       
\graphicspath{{media/}}     
\let\vec\mathbf
\makeatletter
\newcommand{\mathleft}{\@fleqntrue\@mathmargin0pt}
\newcommand{\mathcenter}{\@fleqnfalse}
\makeatother

\newcommand{\beginsupplement}{%
        \setcounter{table}{0}
        \renewcommand{\thetable}{S\arabic{table}}%
        \setcounter{figure}{0}
        \renewcommand{\thefigure}{S\arabic{figure}}%
     }

\newcommand{\rvec}{\left( \vec{I}_{2L} \otimes \vec{v}^T \right) \left( \vec{I}_2 \otimes \vec{\Sigma}_{1:L}  \right) \left( \begin{array}{@{}c@{}} \vec{1}_L \otimes \vec{v} \\ \vec{1}_L \otimes \vec{w} \end{array}  \right) }

\newcommand{\rvecT}{
\left( \begin{array}{@{}c@{}} \vec{1}_L^T \otimes \vec{v}^T \\ \vec{1}_L^T \otimes \vec{w}^T \end{array}  \right)
\left( \vec{I}_2 \otimes \vec{\Sigma}_{1:L}^T  \right) 
\left( \vec{I}_{2L} \otimes \vec{v} \right)  }

\newcommand{\Rvec}{
\left( \begin{array}{@{}c@{}} \vec{1}_2^T \otimes \vec{I}_L \otimes \vec{v}^T  \\ \vec{1}_2^T \otimes \vec{I}_L \otimes \vec{w}^T  \end{array} \right) \left( \vec{I}_2 \otimes \tilde{\vec{\Sigma}} \right) \left( \begin{array}{@{}c@{}c@{}} \vec{I}_L \otimes \vec{v} & \vec{0}  \\ \vec{0} & \vec{I}_L \otimes \vec{w}  \end{array} \right)
}

\newcommand{\qvec} {
\left( \vec{I}_L \otimes \vec{v}^T \right) \vec{\Sigma}_{1:L}  \left( \vec{1}_L \otimes \vec{v} \right)
}

\newcommand{\qvecT} {
 \left( \vec{1}_L^T \otimes \vec{v}^T \right)
 \vec{\Sigma}_{1:L}^T 
\left( \vec{I}_L \otimes \vec{v} \right) 
}

\newcommand{\Qvec}{
\left( \vec{I}_L \otimes \vec{v} \right)^T \tilde{\vec{\Sigma}}  \left( \vec{I}_L \otimes \vec{v} \right)
}

\title{Learning latent causal relationships in \\ multiple time series
}

\author{
  Jacek P. Dmochowski \\
  Department of Biomedical Engineering \\
  City College of New York \\
  New York, NY 10031\\
  \texttt{jdmochowski@ccny.cuny.edu} \\
}

\begin{document}
\maketitle

\begin{abstract}
Identifying the causal structure of systems with multiple dynamic elements is critical to several scientific disciplines. The conventional approach is to conduct statistical tests of causality, for example with Granger Causality, between observed signals that are selected \emph{a priori}. Here it is posited that, in many systems, the causal relations are embedded in a \emph{latent} space that is expressed in the observed data as a linear mixture. A technique for blindly identifying the latent sources is presented: the observations are projected into pairs of components -- driving and driven -- to maximize the strength of causality between the pairs. This leads to an optimization problem with closed form expressions for the objective function and gradient that can be solved with off-the-shelf techniques. After demonstrating proof-of-concept on synthetic data with known latent structure, the technique is applied to recordings from the human brain and historical cryptocurrency prices. In both cases, the approach recovers multiple strong causal relationships that are not evident in the observed data. The proposed technique is unsupervised and can be readily applied to any multiple time series to shed light on the causal relationships underlying the data.
\end{abstract}

\keywords{Granger Causality \and blind source separation \and unsupervised learning}

\section{Introduction}
Several important problems in the sciences are concerned with the activity of interacting sources. Two prominent examples are the dynamics of the brain \cite{sporns2010networks}, where information processing manifests as a spatiotemporal pattern of regional activations constrained by anatomical and functional connectivity, and financial markets \cite{tsay2005analysis}, where values of assets evolve in concert with the decisions of agents. In complex systems such as these, it is critical to infer the underlying structure governing system evolution, and relatedly, to forecast future outcomes.

Granger Causality \cite{granger1969investigating} is a popular technique for measuring a form of dependence rooted in the temporal precedence effect. Time series $x(t)$ is said to cause, in a Granger sense, time series $y(t)$ if the past of $x$ improves the prediction of the present value of $y$ above that of its own past.
Originating in economics\cite{granger2001essays}, Granger Causality has since found extensive utilization in neuroscience \cite{ding200617,seth2015granger}, where it has been applied to recordings of brain activity captured at various spatial and temporal scales to illuminate neural circuits \cite{ding200617,bernasconi1999directionality,kaminski2001evaluating,goebel2003investigating,sheikhattar2018extracting,vicente2011transfer}. Perhaps driven by the ubiquitous interest in causal interactions, the technique has been adopted by many disparate fields, including ecology \cite{sugihara2012detecting}, 
computational biology \cite{finkle2018windowed}, 
and epidemiology \cite{eichler2010granger,kleinberg2011review}.  The utility of Granger Causality has been aided by several extensions and reformulations of the original technique, most notably a frequency-domain formulation \cite{geweke1982measurement,geweke1984measures} and a generalization to multivariate time series \cite{barrett2010multivariate,barnett2014mvgc}. Moreover, several approaches to capturing non-linear causal interactions between multiple time series have been proposed \cite{hiemstra1994testing,ancona2004radial,marinazzo2008kernel,tank2018neural}.
Conventionally, these different variants of Granger Causality are measured between observed signals that are selected \emph{a priori}. In other words, one must specify the identity of the signals being probed, and the hypothesized direction of causality. Moreover, this approach implicitly assumes that the underlying causal relationships exist in the native space defined by the observations (e.g. the sensors). 

The central idea proposed here is that, in many systems, the true causal relations are embedded in a \emph{latent} source space, and that these latent sources enter the observations via an unknown linear mixture. Due to the mixing process, direct application of tools such as Granger Causality to the observed data may not optimally reveal the dynamics of the system. Rather, the approach taken here to identify the latent causal sources is to project the observations into a component space that maximizes the Granger Causality among \emph{pairs} of time series: one signal models the ``driving'' source, and the other captures the source being ``driven''. It is shown that this can be formulated as a non-convex optimization problem with closed-form expressions for the objective function and gradient.  Importantly, the optimization does not require access to the mixing process and thus constitutes blind identification. 

To solve the optimization problem, a simple coordinate descent algorithm that is implemented with standard numerical packages is presented. By simulating a vector autoregressive (VAR) system with known structure, it is demonstrated that the proposed technique indeed identifies the underlying sources, their connections, and the mixing process. To evaluate the proposed approach on real-world systems, the technique is then applied to data from the human brain and the cryptocurrency market. In both cases, it is shown that the proposed technique recovers multiple pairs of signals whose causal strength is significantly greater than what is found in the observed data. 

The distinctions between Granger and true physical causality have been previously described \cite{maziarz2015review,grassmann2020new}. In what follows, the terms ``causal'' and ``causality'' are employed for conciseness with the understanding that the findings presented here pertain to the Granger form of causality.

\section*{Results}

\subsection*{Motivating example}
Consider a simple system with two connected sources, $s_1$ and $s_2$, where source 1 ``Granger causes'' source 2, denoted here by $s_1 \rightarrow s_2$. In the neuroscience context, $s_1$ may represent the mass synaptic activity at a brain region, and $s_2$ the activity of a downstream region to which $s_1$ projects. Due to signal mixing (e.g. volume conduction), the observed signals are modeled as a linear mixture of the two sources:
\begin{eqnarray}
\label{eqn:ill}
    \left( \begin{array}{cc} x_1(t) \\ x_2 (t) \end{array} \right) = 
    \left( \begin{array}{cc} A_{11} & A_{12} \\ A_{21} & A_{22} \end{array} \right) \left( \begin{array}{cc} s_1(t) \\ s_2 (t) \end{array} \right),  
\end{eqnarray}
where the 2-by-2 mixing matrix is assumed to be invertible and where sensor noise has been omitted for the sake of this illustrative example. Note that $s_1$ and $s_2$ are mixed together in the captured signals, potentially confounding the measure of Granger Causality between $x_1$ and $x_2$.  Given only the observations, the goal is to identify the ``driving'' signal $y(t)\approx s_1(t)$ and the ``driven'' signal $z(t)\approx s_2(t)$. Writing (\ref{eqn:ill}) in matrix notation as $\vec{x}(t) = \vec{A} \vec{s} (t)$, $s_1$ is exactly recovered if $y(t)={\vec{w}^{*}}^T \vec{x}(t)$, where $\vec{w}^{\ast}$ is a column vector whose elements are the first row of $\vec{A}^{-1}$ and $^T$ is the transpose operation. Similarly, $s_2$ is recovered as $z(t)= {\vec{v}^{*}}^T \vec{x}(t)$, where $\vec{v}^{\ast}$ is the second row of $\vec{A}^{-1}$. The projection vectors ${\vec{w}^{*}}$ and ${\vec{v}^{*}}$ undo the mixing process by combining the observed signals to form latent components that approximate the underlying sources. The problem considered here is whether it is possible to recover $s_1$ and $s_2$ \emph{without} access to the mixing process $\vec{A}$. Below, a novel criterion for blind source separation that maximizes the Granger Causality between pairs of component signals is proposed.   

\subsection*{Maximizing latent Granger Causality}
Given an observable, centered random process ${\bf x}(t) \in \mathbb{R}^D$, the goal is to identify latent variables $y(t)={\bf w}^T {\bf x}(t)$ and $z(t)={ \bf v}^T {\bf x}(t)$ such that the Granger Causality from $y$ to $z$ is maximized. Namely, it is desired to solve the following optimization problem:
\begin{equation}
\label{eqn:GCAopt}
    \max_{{\bf w} , {\bf v}} \mathcal{G}_{y \rightarrow z}
\end{equation}
where 
\begin{equation}
\label{eqn:GCdef}
    \mathcal{G}_{y \rightarrow z} = 1 - \frac{ E \{ \epsilon_f^2 \} }{E \{ \epsilon_r^2 \}} 
\end{equation}
is termed the ``strength of causality'' \cite{granger1969investigating} from $y$ to $z$, $\epsilon_f$ is the residual of a linear regression predicting $z$ from the history of both $z$ \emph{and} $y$ (i.e., the ``full'' model), and $\epsilon_r$ is the residual when regressing $z$ onto only its past (the ``reduced'' model). $\mathcal{G}_{y \rightarrow z} $ is bounded between 0 and 1, with $\mathcal{G}_{y \rightarrow z}=0$ indicating that $y$ does not aid in the prediction of $z$, and $\mathcal{G}_{y \rightarrow z}=1$ denoting a zero-error estimate of $z$ from the past of itself and $y$. The optimization in (\ref{eqn:GCAopt}) is aimed at identifying two projection vectors, $\vec{w}\in \mathbb{R}^D$ and $\vec{v}\in \mathbb{R}^D$, such that the resulting pair of latent variables maximize the strength of causality (\ref{eqn:GCdef}).

In what follows, a history of $L$ samples is assumed, and the temporal apertures of $y$ and $z$ are defined as:
\begin{eqnarray*}
    \vec{y}_p &=& \left( \begin{array}{ccc} y(t-1) & \ldots & y(t-L) \end{array} \right)^T  \\
    \vec{z}_p &=& \left( \begin{array}{ccc} z(t-1) & \ldots & z(t-L) \end{array} \right)^T. 
\end{eqnarray*}
To arrive at a form of (\ref{eqn:GCdef}) that can be optimized using gradient-based techniques, note that the minimum mean squared error (MMSE) corresponding to the full and reduced models are given by \cite{wiener1964extrapolation}:
\begin{eqnarray}
\label{eqn:Phif_Phir}
    &&\Phi_f = E \{ \epsilon_f^2 \} = \sigma_z^2 - \vec{r}^T \vec{R}^{-1}\vec{r} \nonumber \\
    &&\Phi_r = E \{ \epsilon_r^2 \} =\sigma_z^2 - \vec{q}^T \vec{Q}^{-1}\vec{q},
\end{eqnarray}
where $\sigma_z^2 = E \left\{ z^2(t)  \right\}$ is the mean power of $z$, $E \left\{ \cdot \right\}$ denotes mathematical expectation,
\begin{eqnarray*}
  \vec{r} = E \left\{ z(t) \left( \begin{array}{c} \vec{z}_p(t) \\ \vec{y}_p(t) \end{array}  \right) \right\}  
   ~~~~~~~
     \vec{q}=E \left\{ z(t)  \vec{z}_p(t)  \right\}
\end{eqnarray*}
are $2L$ and $L$ dimensional covariance vectors between $z$ and the temporal apertures of the full and reduced models, respectively, and where
\begin{eqnarray*}
     \vec{R} = E \left\{ \left( \begin{array}{c} \vec{z}_p(t) \\ \vec{y}_p(t) \end{array}  \right) \left( \begin{array}{c} \vec{z}_p(t) \\ \vec{y}_p(t) \end{array}  \right)^T      \right\} 
     ~~~
     \vec{Q} = E \left\{  \vec{z}_p(t)   \vec{z}^T_p(t)        \right\} 
\end{eqnarray*}
are $2L$-by-$2L$ and $L$-by-$L$ covariance matrices of the predictors in the full and reduced models, respectively. Importantly, $\sigma_z^2=\vec{v}^T \vec{\Sigma}(0) \vec{v}$, $\vec{r}$, $\vec{q}$, $\vec{R}$ and $\vec{Q}$ can each be expressed in terms of the projection vectors $\vec{w}$ and $\vec{v}$ and the spatiotemporal statistics of the observations (see \emph{Supplementary Note 1}):
\begin{eqnarray}
\label{eqn:qrRq_vw}
 && \vec{r}= \left( \vec{I}_{2L} \otimes \vec{v}^T \right) \left( \vec{I}_2 \otimes \vec{\Sigma}_{1:L}  \right) \left( \begin{array}{c} \vec{1}_L \otimes \vec{v} \\ \vec{1}_L \otimes \vec{w} \end{array}  \right) 
    ~~~~~~~~~~~~~~~~~~~~~~~~~~~~~~~~~~~ \vec{q}= \left( \vec{I}_L \otimes \vec{v}^T \right) \vec{\Sigma}_{1:L}  \left( \vec{1}_L \otimes \vec{v} \right) \nonumber \\
    &&\vec{R}=\left( \begin{array}{c} \vec{1}_2^T \otimes \vec{I}_L \otimes \vec{v}^T  \\ \vec{1}_2^T \otimes \vec{I}_L \otimes \vec{w}^T  \end{array} \right) \left( \vec{I}_2 \otimes \tilde{\vec{\Sigma}} \right) \left( \begin{array}{cc} \vec{I}_L \otimes \vec{v} & \vec{0}  \\ \vec{0} & \vec{I}_L \otimes \vec{w}  \end{array} \right)  
    ~~~~~~~ \vec{Q}= \left( \vec{I}_L \otimes \vec{v} \right)^T \tilde{\vec{\Sigma}}  \left( \vec{I}_L \otimes \vec{v} \right), 
\end{eqnarray}
where 
\begin{eqnarray*}
    \vec{\Sigma}_{1:L} = \left( \begin{array}{cccc} \vec{\Sigma}(1) & \vec{0} & \ldots  &  \vec{0} \\ 
    \vec{0} & \vec{\Sigma}(2) & \ldots & \vec{0} \\ 
    \vdots & \vec{0} & \ddots & \vdots \\ 
    \vec{0} & \ldots & \ldots & \vec{\Sigma}(L)  \\ 
    \end{array} \right) 
\end{eqnarray*}
is an $LD$-by-$LD$ block covariance matrix where $\vec{\Sigma}(\tau)= E \left\{ \vec{x}(t) \vec{x}^T(t-\tau) \right\}$ is the lagged covariance of the observations, 
\begin{eqnarray*}
    \tilde{\vec{\Sigma}} = \left( \begin{array}{cccc} \vec{\Sigma}(0) & \vec{\Sigma}(-1) & \ldots  &  \vec{\Sigma}(-L+1) \\ 
    \vec{\Sigma}(1) & \vec{\Sigma}(0) & \ldots & \vec{\Sigma}(-L+2)\\ 
    \vdots & \vec{\Sigma}(1) & \ddots & \vdots \\ 
    \vec{\Sigma}(L-1) & \ldots & \ldots & \vec{\Sigma}(0)  \\ 
    \end{array} \right)
\end{eqnarray*}
is an $LD$-by-$LD$ block Toeplitz matrix, $\otimes$ denotes the Kronecker product, $\vec{1}_K$ is a column vector of $K$ ones, and $\vec{I}_K$ is the $K$-by-$K$ identity matrix. Substituting (\ref{eqn:qrRq_vw}) into (\ref{eqn:Phif_Phir}) and the resulting expressions into (\ref{eqn:GCdef}), one arrives at the following expression for the strength of causality between latent sources $y$ and $z$: 
\scriptsize{
\begin{align} 
 {}&   \mathcal{G}_{y \rightarrow z} = 1 - \nonumber \\
 {}&    \frac{\vec{v}^T \vec{\Sigma}(0) \vec{v} - \rvecT \left[ \Rvec \right]^{-1} \rvec }{\vec{v}^T \vec{\Sigma}(0) \vec{v} - \qvecT \left[ \Qvec \right]^{-1} \qvec} \label{eqn:GCdef_vw}
\end{align}
}
\normalsize

The gradient of (\ref{eqn:GCdef_vw}) has a closed-form that is derived in \emph{Supplementary Note 2}. Conventional optimization tools may then be employed to learn projection vectors $\vec{w}^{\ast}$ and $\vec{v}^{\ast}$ that maximize the Granger Causality between resulting latent signals $y(t)={\vec{w}^{\ast}}^T \vec{x}(t)$ and $z(t)={\vec{v}^{\ast}}^T \vec{x}(t)$.

\subsection*{Identifying latent causal structure}
The objective in (\ref{eqn:GCdef_vw}) is non-convex, since $ \mathcal{G}_{y \rightarrow z} \left( \vec{w}, \vec{v} \right) = \mathcal{G}_{y \rightarrow z} \left( a \vec{w}, b \vec{v} \right)$ for arbitrary real scalars $a$ and $b$. This follows from the fact that the residual error when predicting $z$ from $y$ is equivalent to that when predicting $bz$ from $ay$ -- any scaling factors will be accommodated by the temporal filter that predicts the driven signal from the driving signal.  Thus, the technique presented here is only able to identify the latent sources up to a scaling factor. As with other blind source separation techniques such as independent components analysis (ICA) \cite{comon1994independent,hyvarinen2000independent}, it is not possible to recover the scale or sign of the underlying sources.

Another potential ambiguity when optimizing (\ref{eqn:GCdef_vw}) is related to a known property of multivariate Granger Causality \cite{barrett2010multivariate,barnett2014mvgc}. Namely, the strength of causality between $y$ and $z$ is invariant to mixtures of $y$ and $z$ in the driving signal, such that $\mathcal{G}_{y \rightarrow z} = \mathcal{G}_{a y + b z\rightarrow c z}$. This means that, without appropriate modifications to the objective function, maximizing (\ref{eqn:GCdef_vw}) will only identify $z$. To resolve this ambiguity, one can utilize the concept of time-reversed Granger Causality \cite{haufe2013critical,winkler2016validity}. Notice that if $y \rightarrow z$ in $\vec{x}(t)$, then $z \rightarrow y$ in $\vec{x}(-t)$.  Thus, while the ambiguity in forward time occurs on $y$, it occurs in reversed time on $z$. One can therefore combine forward and reversed time into a single objective function according to:
\begin{equation}
\label{eqn:GCAopt_tr}
    \max_{{\bf w} , {\bf v}} \mathcal{G}_{y \rightarrow z} + \mathcal{G}_{z \rightarrow y}^{\mathrm{tr}}, 
\end{equation}
where $\mathcal{G}_{z \rightarrow y}^{\mathrm{tr}}$ is the strength of causality between $z(-t)$ and $y(-t)$. 

The non-convexity of the objective function means that a local minimizer of (\ref{eqn:GCAopt_tr}) is not guaranteed to represent a global minimum. Many approaches to non-convex optimization have been developed, including the use of multiple starting points \cite{ugray2007scatter} and stochastic gradient descent \cite{bottou2018optimization}. Here, a grouped coordinate descent algorithm \cite{bezdek1987local} that maximizes $\vec{v}$ and $\vec{w}$ in an alternating fashion is proposed: instead of combining $\vec{v}$ and $\vec{w}$ into a single model parameter and performing a 2$D$ dimensional optimization, the driving and driven signals are learned in tandem. This reduces the dimensionality of the problem, partitions the variables in a natural manner, and is shown empirically to converge to optima that recover the causal structure underlying the data.  

As the cost function is non-convex, there are potentially several pairs of projection vectors 
that locally maximize the strength of causality (\ref{eqn:GCAopt_tr}) and thus may yield meaningful latent sources. In order to recover $P$ pairs of components $\{y_i(t), z_i(t)\}, i=1,\ldots,P$, here it is proposed to repeat the optimization after the first iterate, but not before removing the contribution of the driving signal $y_1(t)$ from the observed data. This takes the form of a spatiotemporal regression such that any signals that are correlated with $y_1(t)$ or its lagged versions $y_1(t-l), l=1,\ldots,L$ are removed. Given that this should include $z_1(t)$, the driven signal is not explicitly removed. This procedure is repeated until the desired number of component pairs $P$ is obtained. The proposed algorithm is described in \emph{Supplementary Note 3}. 

In what follows, the proposed approach is evaluated on synthetic and real-world data. The primary criterion employed to assess performance is the strength of causality (\ref{eqn:GCdef}) among the recovered pairs of components relative to the strength of causality between observed signals, or those formed by conventional component analysis techniques. Where possible, the fidelity of the recovered signals compared to the underlying sources is measured. Moreover, the recovered components and associated projection vectors are interpreted based on what is known about the system being investigated (i.e., neural dynamics, the cryptocurrency market) to further assess the behavior of the proposed approach.

\subsection*{Recovering the causal structure of a three-element system}
To test the proposed method's ability to recover the causal structure embedded in multiple time series, a series of empirical evaluations was conducted on synthetic data. Access to the system's ground-truth structure permitted measuring the fidelity of the recovered signals with respect to the latent sources. The data was generated according to a VAR(3) process whose parameters matched those employed by Stokes and Purdon \cite{stokes2017study}, where $s_1 \rightarrow s_2$ and $s_2 \rightarrow s_3$. Projection of the three connected  sources to a $D=4$ dimensional observation vector followed as $\vec{x}(t)= \vec{A} \vec{s}(t)$, where the elements of 4-by-3 mixing matrix $\vec{A}$ were randomly drawn from the uniform distribution $A_{ij} \sim U[0,1]$. The proposed technique was employed to recover $P=2$ pairs of components.


The latent sources, observed data, and recovered signals of a single realization are depicted in Fig \ref{fig:sim_var}a,b, and c, respectively. The goal of the proposed approach is to recover the $s_1 \rightarrow s_2$ 
relationship in ($y_1,z_1$), and the $s_2 \rightarrow s_3$ link in ($y_2,z_2$). Notice that $s_2$ is both a driven signal as well as a driving signal, and thus the components $z_1$ and $y_2$ are aiming to capture the \emph{same} signal. The strength of causality among all pairs of latent sources is illustrated in Fig \ref{fig:sim_var}d, where the order dependence inherent to Granger Causality is evident in the asymmetry of the matrix (rows correspond to the driving signals, columns to the driven). The underlying strengths of causality were measured as: $ \mathcal{G}_{s_1 \rightarrow s_2} = 0.11 \pm 0.0011$ and $\mathcal{G}_{s_2 \rightarrow s_3} = 0.10 \pm 0.0010$ (mean $\pm$ sem across $n=100$ random realizations). 

The strength of causality measured among pairs of observed signals was markedly lower, with a maximum (across all pairs) strength of causality of $0.066 \pm 0.002$, significantly lower than the underlying latent causality ($p=3.9 \times 10^{-18}$ comparing to $\mathcal{G}_{s_1 \rightarrow s_2}$, $p=4.0 \times 10^{-18}$ comparing to $\mathcal{G}_{s_2 \rightarrow s_3}$, Wilcoxon signed rank test, $n=100$). The strengths of causality between observed signals is depicted for a single realization in Fig \ref{fig:sim_var}e, where the strongest connection was $\mathcal{G}_{x_1 \rightarrow x_3}=0.025$. Notice that the system structure (i.e., two connected pairs) is no longer apparent, as the mixing process has both obscured and dampened the underlying causal relationships. 

The causality matrix of the $P=2$ recovered components is shown in Fig \ref{fig:sim_var}f (rows and columns correspond to driving and driven components, respectively).  Two strong connections are readily apparent: $y_1 \rightarrow z_1$ and $y_2 \rightarrow z_2$. The magnitudes of these causal relationships closely matched those of the latent sources: $\mathcal{G}_{y_1 \rightarrow z_1} = 0.11 \pm 0.0014$ and $\mathcal{G}_{y_2 \rightarrow z_2} = 0.10 \pm 0.0010$. These values were significantly larger than the maximum causality among all pairs of observed variables (pair 1: $p=4.7 \times 10^{-18}$, pair 2: $p=4.0 \times 10^{-18}$). To determine whether the recovered components captured the underlying sources, the mixing matrix was estimated by regressing the driving and driven signals  onto the observations. The true mixing matrix is depicted for a single realization in Fig \ref{fig:sim_var}g. The recovered matrices exhibited a large correlation with the true values ($r^2=0.98 \pm 0.004$, shown for a single realization in Fig \ref{fig:sim_var}h). Moreover, the time series of recovered components faithfully tracked the dynamics of the latent sources: $r^2=0.98 \pm 0.007$ between $s_1$ and $y_1$ (Fig \ref{fig:sim_var}i), $r^2=0.96 \pm 0.014$ between $s_2$ and $z_1$ (Fig \ref{fig:sim_var}j), $r^2=0.98 \pm 0.004$ between $s_2$ and $y_2$ (Fig \ref{fig:sim_var}k), and $r^2=0.99 \pm 0.002$ between $s_3$ and $z_2$ (Fig \ref{fig:sim_var}l). Note that latent source $s_2$ was indeed captured by both $z_1$ and $y_2$.

\subsection*{Identifying latent causal connections in the brain}
\begin{figure*}[ht!]
\centering\includegraphics[width=\linewidth]{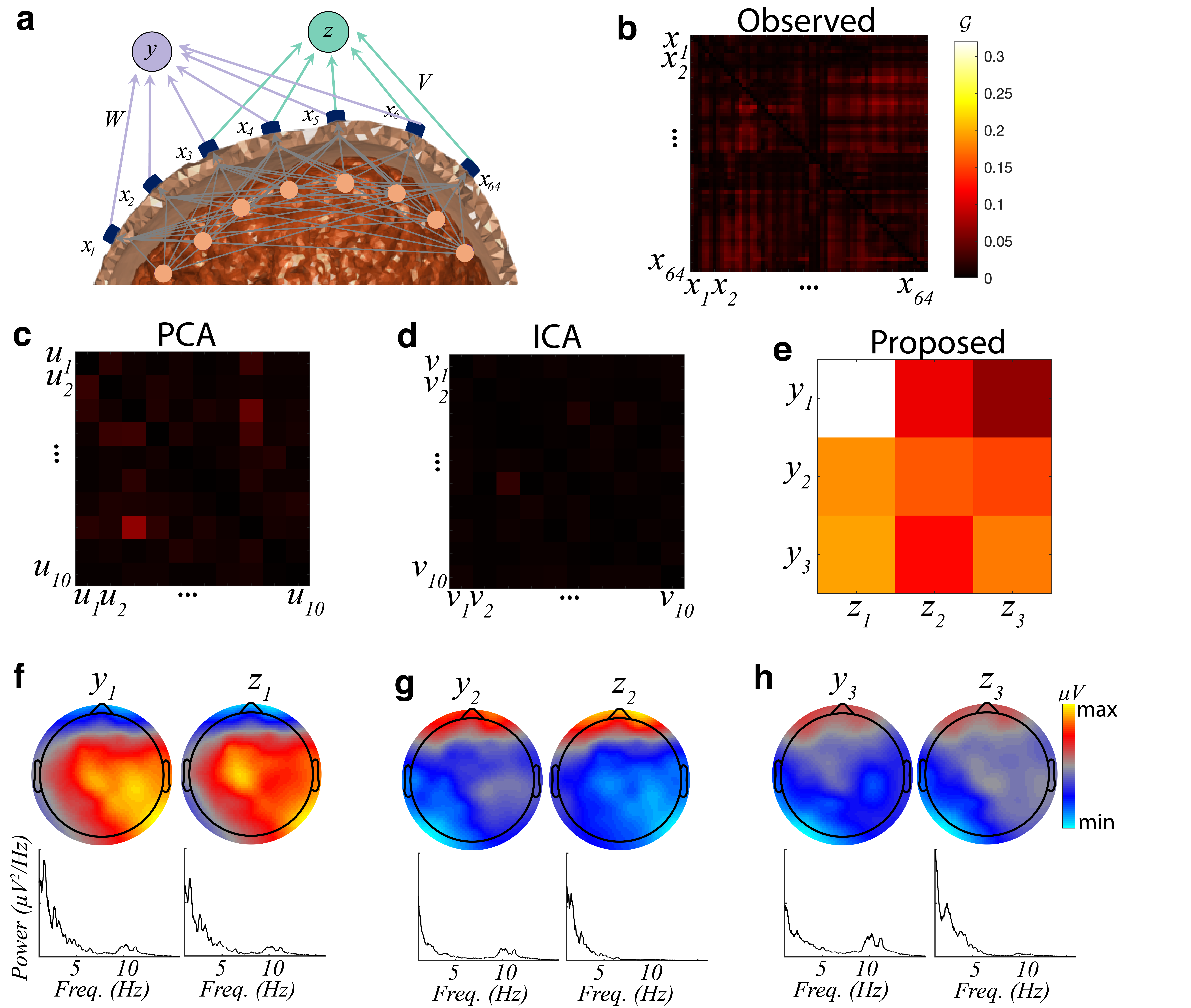}
\caption{ \label{fig:eeg} \textbf{Learning causal relationships in the human brain.} A data set consisting of the scalp EEG of $n=12$ subjects viewing television advertisements was employed to evaluate the causal relationships recovered by the proposed method with those from conventional approaches. 
{\bf (a)} The scalp EEG manifests as a linear mixture of interacting electrical sources in the cerebral cortex. 
{\bf (b)} The strength of causality between all pairs of electrode signals, where the largest value $0.073$ was observed between electrodes CP1 (left centro-parietal) and CP4 (right centro-parietal).
{\bf (c)} The matrix of Granger Causality among the first 10 principal components, where a maximum of $0.067$ was detected between PCs 8 and 3.
{\bf (d)} Same as (c) but now shown for the first 10 independent components, where a maximum of $0.022$ was found between components 6 and 3. 
{\bf (e)} The proposed approach recovered three pairs of components whose strength of causality was $0.32$, $0.16$, and $0.18$, respectively, depicted in the diagonal elements of the matrix. 
{\bf (f)} The scalp topographies (top row) and power spectra (bottom row) of the driving and driven signal in the first pair of recovered components. The driving signal exhibited a right temporo-parietal distribution and large delta (1-3 Hz) power, while the driven signal showed peak expression over the left central electrodes.
{\bf (g)} Same as (f) but now for the second pair of recovered components, where the driving signal originated over the left parieto-occipital region and the driven signal over the right temporo-parietal region. Notice also that the driving and driven signals showed opposing pattterns in their power spectrum: $y_2$ was marked by low delta and high alpha (8-13 Hz)  power, while $z_2$ showed the opposite (high delta and low alpha).
{\bf (h)} The scalp topographies and power spectra of the third pair of recovered components. The driving signal showed bilateral expression in the parieto-occipital region, while the driven signal was concentrated over the left occipital region. Once again, the driving signal showed a high level of alpha band power, and high delta power was observed in the power spectrum of the driven signal.
}
\end{figure*}

Scalp electroencephalogram (EEG) signals, sometimes referred to as ``brain waves'', arise from the coordinated activity of a large number of neurons in the cerebral cortex \cite{buzsaki2012origin}. At any time instant, the set of electric potentials registered by scalp electrodes is a linear mixture of dipolar generators \cite{baillet2001electromagnetic} (Fig \ref{fig:eeg}a). It was hypothesized that Granger causal relations would be most strongly observed at the level of these neural generators, estimated by spatially filtering the EEG \cite{parra2005recipes}. To test this, the proposed technique was applied to a previously collected EEG data set where $n=12$ healthy participants viewed television advertisements that originally aired during the 2012 and 2013 \emph{Super Bowl} football matches \cite{dmochowski2014audience}. 

To identify the level of Granger causality among the captured $D=64$ signals, the strength of causality was measured for all pairs of electrodes (Fig \ref{fig:eeg}b). The strongest relationship was found between left centroparietal electrode ``CP1'' and right centroparietal electrode ``CP4'', with $\mathcal{G}=0.073$ (Fig \ref{fig:eeg}b). In order to determine whether conventional spatial filtering approaches recover stronger causal relationships than those found among electrodes, the observed data was decomposed with both principal components analysis (PCA) and independent components analysis (ICA). Surprisingly, the strength of causality among pairs of PCs and ICs was not larger than that found in the raw electrodes: a maximum value of $\mathcal{G}=0.067$ was found between principal components 8 and 3 (Fig \ref{fig:eeg}c), and a maximum of $\mathcal{G}=0.022$ between independent components 6 and 3 (Fig \ref{fig:eeg}d). Next, the proposed method was employed to recover $P=3$ pairs of latent components. The strength of causality among the recovered pairs was substantially larger, with $\mathcal{G}_{y_1 \rightarrow z_1} = 0.32$, $\mathcal{G}_{y_2 \rightarrow z_2} = 0.16$, and $\mathcal{G}_{y_3 \rightarrow z_3} = 0.18$, for pairs 1, 2, and 3, respectively (Fig \ref{fig:eeg}e). The presence of more than two-fold increases in the strength of causality at multiple component pairs is consistent with the notion that the underlying causal relationships occur in a latent subspace of the data. 

The coefficients of the spatial filter weights learned by the proposed method represent the scalp regions expressing the driving and driven signals. 
For pair 1, the causing signal $y_1$ exhibited peak expression over the right temporo-parietal region, while the driven signal $z_1$ had peak expression over the left central electrodes (Fig \ref{fig:eeg}f). This indicates that, during this task, activity over the right temporo-parietal cortex temporally preceded activity over the left central region. To further interpret the learned components, the power spectrum of the driving and driven signals were measured. The power spectrum of scalp EEG is typically segregated into distinct frequency bands, with a large body of literature documenting associations between cognitive states and activity in specific bands \cite{klimesch1999eeg}. Both $y_1$ and $z_1$ showed high levels of power in the delta band (1-3 Hz), and moderate levels of alpha band (8-13 Hz) power (Fig \ref{fig:eeg}f). The spatial topographies of the next strongest pair showed peak expression over the left parieto-occipital ($y_2$) and right temporo-parietal regions ($z_2$), indicating inter-hemispheric connectivity (Fig \ref{fig:eeg}g). An interesting pattern arose in the power spectra of the components: the driving signal was marked by low delta power and high alpha power, while the driven signal exhibited the opposite pattern (i.e., high delta power and a notable absence of alpha power). This result is consistent with previous findings of an inverse correlation between alpha and delta waves, hypothesized to arise from thalamocortical inhibition of the brain stem \cite{robinson1999technical,robinson2001brain}. The topography of driving signal $y_3$ exhibited activation over the left occipital and right centro-temporal regions, while the corresponding driven signal $z_3$ was concentrated over the left occipital region (Fig \ref{fig:eeg}h). As observed in pair 2, the driving signal showed a high ratio of alpha-to-delta power, while a low alpha-to-delta ratio was detected in the driven signal. 

\begin{figure}[ht!]
\centering\includegraphics[width=1\linewidth]{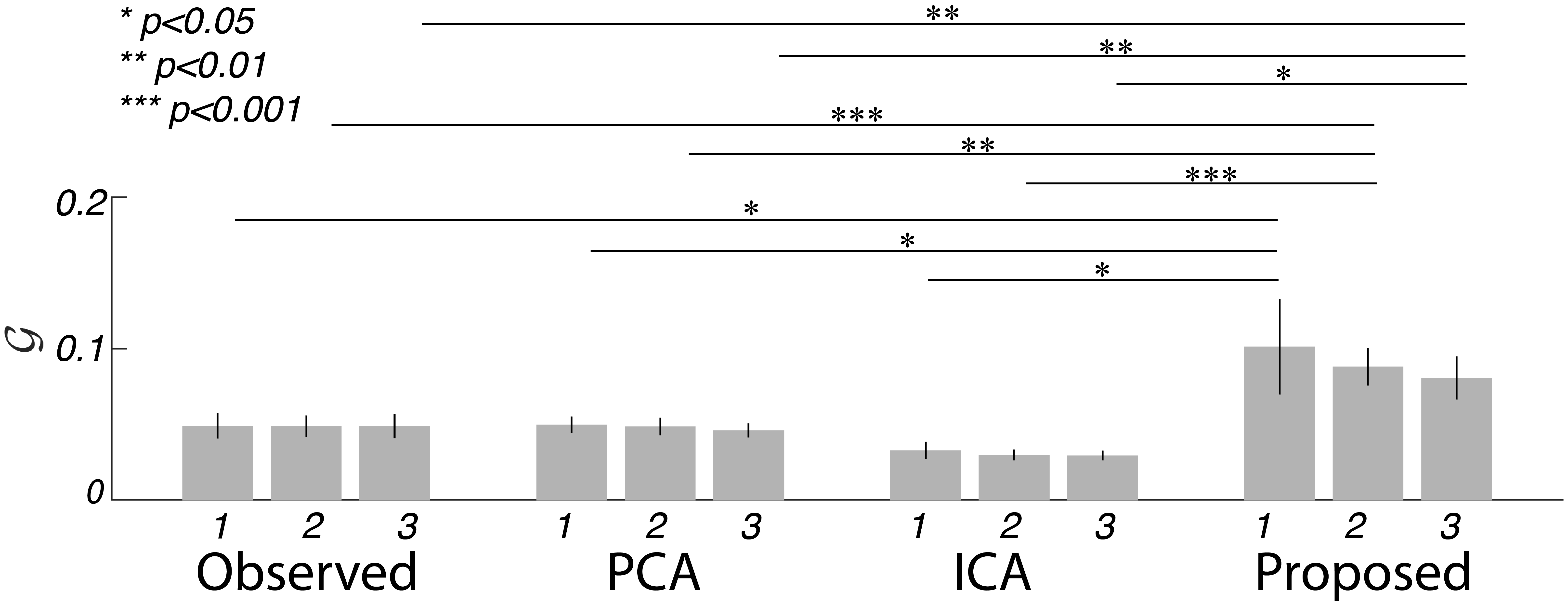}
\caption{ \label{fig:eeg_anova} \textbf{Proposed technique recovers stronger causal relationships than conventional techniques.} EEG data from $n=12$ subjects was employed to test the hypothesis that the component pairs recovered by the proposed approach reflect stronger causal links than yielded by existing approaches.  The strength of causality was measured for the three most causally connected pairs of electrodes, principal components, and independent components (all selected \emph{post hoc}). The resulting values were compared to the strength of causality among the three pairs of recovered components obtained with the proposed approach. A two-way ANOVA with method and component as factors was conducted, yielding a significant main effect of method ($F(3)=11.53$, $p=9.23 \times 10^{-7}$). There was no main effect of component ($p=0.59$), and no interaction ($p=0.98$). Follow-up tests showed that the main effect of method was driven by significantly larger strengths of causality with the proposed method ($\mathcal{G} = 0.10 \pm 0.031$, $0.088 \pm 0.012$, and $0.081 \pm 0.014$ for the first three components, means $\pm$ sem across $n=12$ subjects) relative to the three most connected electrode pairs ($\mathcal{G} = 0.049 \pm 0.0084$, $0.049 \pm 0.0072$, $0.049 \pm 0.0079$; $p=0.034$; $p=4.9 \times 10^{-4}$, and $p=0.034$ for components 1, 2, and 3, respectively; Wilcoxon signed rank test, $n=12$), the three most connected principal component pairs ($\mathcal{G} = 0.050 \pm 0.0054$, $0.049 \pm 0.0058$, $0.046 \pm 0.0046$; $p=0.034$, $p=0.0049$, and $p=0.0093$), and the three most connected independent component pairs ($\mathcal{G} = 0.033 \pm 0.0056$, $0.030 \pm 0.0035$, $0.030 \pm 0.0031$; $p=0.016$, $p=4.9 \times 10^{-4}$, and $p=0.0024$). This provides evidence that the proposed approach is able to detect stronger causal relationships than what is found in the observed data and conventional components analyses. }
\end{figure}

To formally test whether the proposed method recovers stronger causal relations than those found with conventional approaches, a two-way ANOVA (method $\times$ component) was conducted. For this analysis, the strength of causality was measured separately for each subject, yielding $n=12$ repeated measures. A large main effect of method was identified ($F(3)=11.53$, $p=9.23 \times 10^{-7}$; Fig \ref{fig:eeg_anova}). There was no main effect of component ($p=0.59$) and no significant interaction ($p=0.98$). Follow-up tests showed that the main effect of method was driven by significantly larger strengths of causality with the proposed method ($\mathcal{G} = 0.10 \pm 0.031$, $0.088 \pm 0.012$, and $0.081 \pm 0.014$ for the first three components, means $\pm$ sem across $n=12$ subjects) relative to the three most connected electrode pairs ($\mathcal{G} = 0.049 \pm 0.0084$, $0.049 \pm 0.0072$, $0.049 \pm 0.0079$; $p=0.034$; $p=4.9 \times 10^{-4}$, and $p=0.034$ for components 1, 2, and 3, respectively; Wilcoxon signed rank test, $n=12$), the three most connected principal component pairs ($\mathcal{G} = 0.050 \pm 0.0054$, $0.049 \pm 0.0058$, $0.046 \pm 0.0046$; $p=0.034$, $p=0.0049$, and $p=0.0093$), and the three most connected independent component pairs ($\mathcal{G} = 0.033 \pm 0.0056$, $0.030 \pm 0.0035$, $0.030 \pm 0.0031$; $p=0.016$, $p=4.9 \times 10^{-4}$, and $p=0.0024$). Thus, the proposed technique detected causal relationships whose magnitude was significantly larger than those measured with conventional approaches. 

\subsection*{Probing latent causality in the cryptocurrency market}
\begin{figure*}[ht!]
\centering\includegraphics[width=\linewidth]{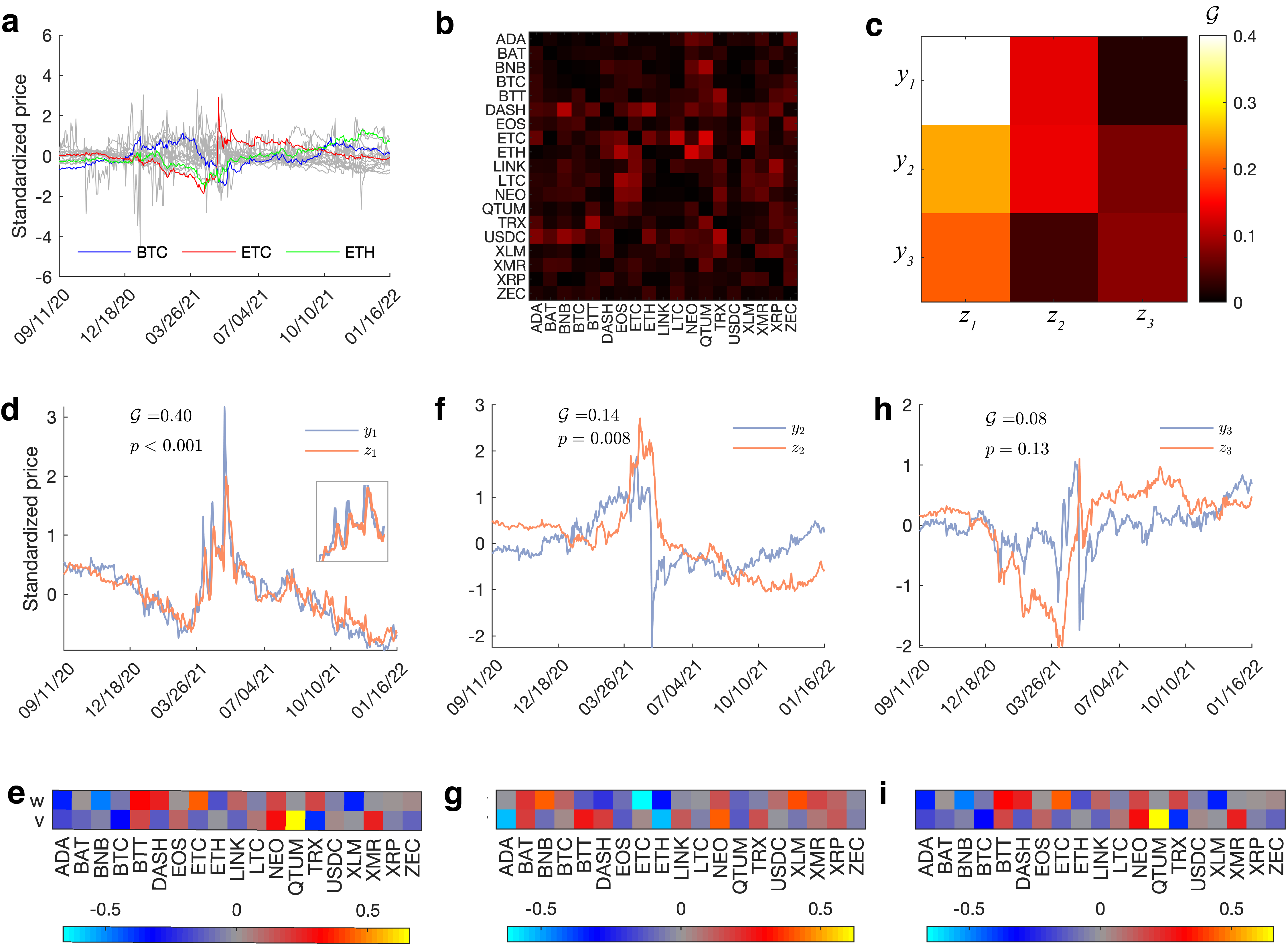}
\caption{ \label{fig:crypto} \textbf{Probing causality in the cryptocurrency market.} {\bf (a)} Historical price data was obtained for 19 cryptocurrencies spanning a time period of approximately 15 months. The individual time series have been standardized to accommodate a large dynamic range between currencies. {\bf (b)} The conventional Granger Causality measure between pairs of individual currencies was modest (mean $\pm$ sd of $0.028 \pm 0.023$), peaking at a value of 0.12 at ETC $\rightarrow$ LTC. {\bf(c)} On the other hand, the proposed method recovered two pairs of latent variables with strong and statistically significant strengths of causality: $\mathcal{G}_{y_1 \rightarrow z_1}=0.40$ ($p<0.001$, non-parametric permutation test with 1000 phase-randomized surrogate data records) and $\mathcal{G}_{y_2 \rightarrow z_2}=0.14$ ($p=0.008$), and a third pair whose strength of causality fell short of significance but still exceeded $95\%$ of individual currency pair values ($\mathcal{G}_{y_3 \rightarrow z_3}=0.080$, $p=0.13$). {\bf(d)} The dynamics of driving signal $y_1$ clearly precede those of the driven signal $z_1$, with the time series peaks consistently occurring a few samples earlier for $y_1$ (see inset). {\bf(d)} The weights of the filter employed to construct the driving signal ($\vec{w}$, top row) and those used to construct the driven signal ($\vec{v}$, bottom row) show that the largest contribution to $y_1$ was exhibited by the currencies BNB and ETC, while QTUM and TRX were most strongly expressed by $z_1$ (note that the magnitudes of the depicted weights represent the contribution of the individual currencies to the latent variables). {\bf(f)} Past values of driving signal $y_2$ can be seen to predict present values of driven signal $z_2$, with the largest inflection of $y_2$ occurring a short time before the corresponding dip in $z_2$. {\bf(g)} Similar to the first latent pair, the currencies ETC and BNB contributed most strongly to $y_2$, but the driven signal here best expressed ADA and ETH. {\bf(h)} The dynamics of the third pair of latent variables exhibited a dynamic distinct from those of the first two pairs, with the temporal precedence of $y_3$ relative to $z_3$ once again visible in the middle region of the curves. {\bf(i)} The cryptocurrencies BNB and XRP most strongly contributed to $y_3$, while $z_3$ was best expressed by XLM and ETC. }
\end{figure*}

Finally, the proposed method was tested on a system without an obvious latent structure: the cryptocurrency market. Historical prices of $D=19$ popular cryptocurrencies (Fig \ref{fig:crypto}A, individual traces have been standardized), were employed for the analysis, which sought to identify the $P=3$ strongest causal relationships. 

Among pairs of individual cryptocurrencies, the strength of causality was quite modest: $0.028 \pm 0.023$ (mean $\pm$ sd across all $n=342$ pairs of currencies), with a maximum value of 
$\mathcal{G}_{ \mathrm{ETC} \rightarrow \mathrm{QTUM}} = 0.12$ (Fig \ref{fig:crypto}B). In contrast, the proposed technique identified a primary pair of latent components with a statistically significant strength of causality ($\mathcal{G}_{y_1 \rightarrow z_1}=0.40$, $p<0.001$, non-parametric permutation test altering the phase of individual cryptocurrency time series), representing a more than three-fold increase (Fig \ref{fig:crypto}C). A statistically significant strength of causality was also found for the second pair of components ($\mathcal{G}_{y_2 \rightarrow z_2}=0.14$, $p=0.008$; Fig \ref{fig:crypto}C). Note that, even after removing the contribution from the primary driving signal $y_1$, a latent relationship whose causality exceeded that seen in the observed data was still recovered. The strength of causality exhibited by the third pair of latent components ($\mathcal{G}_{y_3 \rightarrow z_3}=0.080$, $p=0.13$) fell short of significance, but nevertheless exceeded 96\% of the individual pair values (compare panels B and C in Fig \ref{fig:crypto}).

The dynamics of the driving and driven components of the first pair are depicted in Fig \ref{fig:crypto}D, where the temporal precedence of $y_1$ relative to $z_1$ is visible in the traces. For example, note that the occurrence of the three prominent peaks in the spring of 2021 is first observed in $y_1$ and shortly after in $z_1$ (see Fig \ref{fig:crypto}D inset). The individual currencies with the largest expression in the driving signal were BNB (Binance Coin) and ETC (Ethereum Classic), while the largest contributions to the driven signal were from QTUM and TRX (Fig \ref{fig:crypto}E, color indicates weight of filter used to construct $y_1$ and $z_1$). This result indicates that past fluctuations in the prices of BNB and ETC predict the current prices of QTUM and TRX. The temporal precedence of $y_2$ relative to $z_2$ is also evident in the dynamics of the second pair of latent components (Fig \ref{fig:crypto}F). For example, a sharp dip in price occurs near May 2021, first in $y_2$ and slightly later in $z_2$. Similar to the first pair of latent components, the currencies best expressed in $y_2$ were ETC and BNB. However, unlike ($y_1,z_1$), the driven signal here most strongly expressed ADA (Cardano) and ETH (Ethereum) (Fig \ref{fig:crypto}G). The finding of similar driving signals (but distinct driven signals) in the first two pairs suggests the presence of multiple ``links'' emanating from the latent driver. The currencies best expressed in the driving signal of the third pair were BNB and XRP (Ripple), while the corresponding driven signal $z_3$ best expressed XLM (Stellar) and ETC (Fig \ref{fig:crypto}I). 

\section*{Discussion}

The distinction between the proposed technique and conventional univariate and multivariate Granger Causality can be illuminated by the types of queries that the different approaches address. In the context of the cryptocurrency market, univariate Granger Causality addresses questions such as ``does the price of Bitcoin exert a causal influence on the price of Ethereum?'' Multivariate Granger Causality is concerned with questions such as ``do the prices of Bitcoin and Cardano (taken as a group) drive the prices of Ethereum and Ethereum Classic?'' Note that, in both cases, one must specify the elements and direction of the causal relationship being tested. To identify the full complement of causal links in the system of interest, such a hypothesis testing approach will generally require a large number of statistical tests. In contrast, the proposed method automatically identifies paired groups of cryptocurrencies, with each group defined such that the strength of causality from the driving group to the driven group is maximized: the elements and direction of the causal links are learned directly from the data. This identification may be performed over several iterations, with each iteration revealing a generally weaker but distinct causal relationship from the previous. The weights of the learned filters are interpretable: dimensions with a large magnitude indicate that the corresponding signal is either driving activity, or being driven, in a latent subspace of the system. 

In applications such as EEG or magnetoencephalography (MEG) where the source space has a clear physical substrate, the learned filters offer clear insight into the nature of the latent sources. Namely, the cortical generators of the scalp topographies in Fig \ref{fig:eeg}f-h may be estimated with source localization \cite{baillet2001electromagnetic} to estimate the spatial origin of the latent sources. Causal relationships that are obscured at the level of the electrodes may be clarified as genuine connections between cortical sources. The nature of the latent source space is less apparent in other problems. In financial systems defined by a set of evolving prices, the latent sources correspond to a set of linked assets whose dynamics exhibit a temporal dependence on those of a second set. For example, the occurrence of an external event (e.g. activity on social media) may produce a change in the value of a certain group of assets. As a consequence, the value of a second (disparate) group of assets may also be modulated, and due to the delay between the price movements, a Granger causal relationship emerges. 

Conventional approaches to blind source separation assume that the underlying sources are statistically independent, perhaps inspired by the ``cocktail party problem'' \cite{mcdermott2009cocktail}  solved by the auditory system. This assumption is exploited by Independent Components Analysis (ICA) \cite{comon1994independent,hyvarinen2000independent}, which projects the observed signals into components to maximize their statistical independence. On the other hand, the approach proposed here assumes the existence of Granger Causal sources, and is thus applicable to systems with temporal dependencies  among the signals of interest. Notice that the criteria optimized by ICA and the proposed technique to perform source separation are opposing. In the context of brain signals, ICA is seeking to identify decoupled neural sources, while the method proposed here aims to recover functionally connected brain regions. More closely related to the proposed method are approaches that combine Canonical Correlation Analysis \cite{hotelling1992relations} with Granger Causality \cite{sato2010analyzing,wu2011kernel} to test causal relations between pairs of multivariate time series. These approaches share a feature of the proposed method by forming components of observed data, but differ importantly in that the data must already be partitioned into hypothesized driving and driven signals. 

One limitation of the proposed technique is the potential difficulty in identifying causality in data with very high dimensionality (i.e., the number of observed signals) or very long temporal dependencies between latent sources. In either case, the covariance matrices required to identify the latent causal sources may be poorly estimated, potentially leading to erroneous estimates of latent Granger Causality. To mitigate this, it is required to assume some prior information about the structure of the observed signals. For example, a form of Tikhonov regularization \cite{golub1999tikhonov} equivalent to adding uncorrelated noise to the measurements was employed here. More sophisticated approaches to covariance estimation in high dimensions will improve the performance of the proposed framework. 

A challenge with conventional Granger Causality is the potential presence of exogenous sources that drive two or more observed variables with different delays. In this event, spurious relationships between the observed signals may be inferred. To address this, partial Granger Causality \cite{guo2008partial} may be employed to measure the relationship that remains after removing the contribution of the exogenous source. It is interesting to consider how such confounding sources may affect the behavior of the proposed technique. If the nature of the confounding source is known \emph{a priori}, it should be regressed out of the data prior to deploying the proposed technique. This was performed in the cryptocurrency example above, where the global market trend was removed prior to analysis. In the case of an unknown confounding source, the proposed approach is expected to provide some shielding from spurious inference. This follows from the utilization of multiple component pairs to separate the contributions of distinct latent sources.  For example, in the case of a strong confounding source that enters the observed data, the underlying relationship may appear in the first pair of latent sources, leaving the genuine causal relationships in subsequent pairs. The technique proposed here is tasked with capturing all latent sources that produce Granger Causal links, meaningful or otherwise. This highlights the importance of interpreting the weights of the learned projection vectors, which may offer clues as to the origin of the recovered relationship. 

Granger Causality is one of several statistical approaches to measuring causality. Two popular frameworks that have been successfully applied to dynamic systems are Dynamic Causal Modeling (DCM) \cite{friston2003dynamic} and Structural Equation Modeling \cite{mcintosh1991structural}. In DCM, a ``forward model'' that relates the activity of underlying sources to the observations is specified, with Bayesian model selection utilized to estimate the parameters of the underlying sources (i.e., connectivity). This allows DCM to take advantage of the known structure of the system, including nonlinear interactions. The approach proposed here, while also aiming to identify causal structure, is complementary in nature. The forward model need not be specified beforehand, and the technique functions not as a statistical test \emph{per se} but rather a decomposition of the data, akin to PCA and ICA. Moreover, the knowledge gleaned from the components recovered by the decomposition may then be employed in a subsequent hypothesis testing procedure that has been informed by the method's findings.

\section*{Materials and Methods}
All data and source code are provided at \href{dmochow.github.io/gca}{\fontfamily{pcr}\selectfont dmochow.github.io/gca}. Data analysis was performed in the MATLAB computing environment (Mathworks, Natick MA). 


\paragraph{Implementation} To solve the optimization problems at each iteration of the grouped coordinate descent algorithm (see Algorithm 1 in \textit{Supplementary Note 3}), we employed the built-in MATLAB function {\fontfamily{pcr}\selectfont
fmincon} with the default interior point algorithm solver. The maximum number of function evaluations was set to $10^{4}$ and the maximum number of iterations was set to $4000$. Regularization of the block covariance matrices $\vec{\Sigma}_{1:L}$ and $\tilde{\vec{\Sigma}}$ was implemented by limiting the condition number of each matrix to a value of $c$, where the value of $c$ was selected based on the dimensionality of the problem, as specified below. Limiting the condition number was implemented by adding a small diagonal component $\sigma^2 \vec{I}$ to each covariance matrix, where the value of $\sigma^2 = \frac{ ( \lambda_1 - \lambda_{LD} c) }{c-1}$ ensures that the condition number of the covariance matrix is $c$, where $\lambda_1$ and $\lambda_{LD}$ are the largest and smallest eigenvalues of the block covariance matrix being regularized \cite{hoerl1970ridge,tabeart2020improving}.   


Although the closed-form expression for the gradient of $\mathcal{G}_{y \rightarrow z}$ (see \emph{Supplementary Note 2}) was verified empirically, it was more efficient to compute the gradient numerically with finite differences. The numerous Kronecker products and matrix inverse operations required to evaluate the gradient expression led to longer run times compared to the finite differences approximation. Moreover, in order to guarantee that the optimization identified projections with unit norm, a pair of nonlinear constraints were added, leading to the following constrained optimization problem:
\begin{eqnarray}
    \min_{\vec{w},\vec{v}} ~ -\left[  \mathcal{G}(\vec{w},\vec{v}) + \mathcal{G}^{\mathrm{tr}}(\vec{v},\vec{w}) \right] \nonumber  \\  \mathrm{~~~subject~to:~}  \vec{w}^T \vec{w}=1 \mathrm{~and~}  \vec{v}^T \vec{v}=1,
\end{eqnarray}
where 
$\mathcal{G}(\vec{w},\vec{v})$ is the strength of causality (\ref{eqn:GCdef}) between driving signal $\vec{w}^T \vec{x}(t)$ and driven signal $\vec{v}^T \vec{x}(t)$, and $\mathcal{G}^{\mathrm{tr}}(\vec{v},\vec{w})$ is the  strength of causality between driving signal $\vec{v}^T \vec{x}(-t)$ and driven signal $\vec{w}^T \vec{x}(-t)$. After each iteration of the grouped coodinate descent, the driving signal $y(t)={ \vec{w}^{\ast} }^{T} \vec{x}(t)$ and its lagged versions were regressed out of the data according to:
\begin{eqnarray}
    \vec{x}(t) &=& \vec{x}(t) - \vec{B}^T \vec{y}_p(t) 
\end{eqnarray}
where $\vec{B} = \vec{Y}_p^{\#} \vec{X}$ is the least-squares solution to the linear system:
\begin{eqnarray}
    \vec{X} = \vec{Y}_p \vec{B}
\end{eqnarray}
where $D$-by-$T$ matrix $\vec{X} = \left[ \begin{array}{ccc} \vec{x}(1) & \ldots & \vec{x}(T) \end{array} \right] $ and $L$-by-$T$ matrix  $ \vec{Y}_p = \left[ \begin{array}{ccc} \vec{y}_p(1) & \ldots & \vec{y}_p(T) \end{array} \right] $ span the spatiotemporal apertures of the observed and driving signals, respectively. Convergence was assessed after every iteration, and the search was stopped when the magnitude of change in both $\mathcal{G}$ and $\mathcal{G}^{\mathrm{tr}}$ was less than $10^{-6}$.

\begin{figure}[ht!]
\centering\includegraphics[width=1\linewidth]{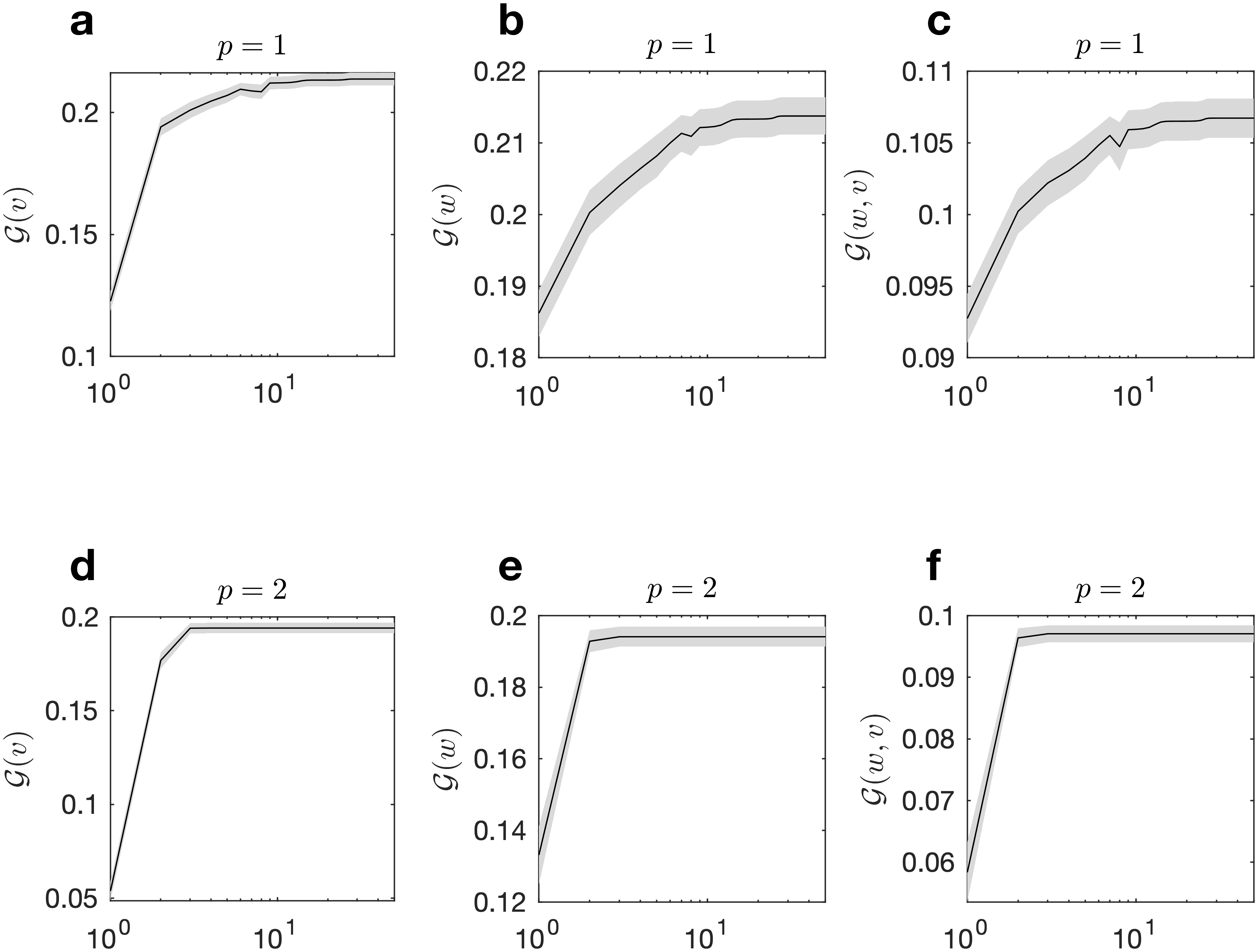}
\caption{ \label{fig:convergence} \textbf{Convergence of grouped coordinate descent on VAR(3) process data.} Vertical axes depict the objective function of the proposed approach, namely the strength of causality between the driving and driven signal. Horizontal axes depict the iteration number in logarithmic spacing. {\bf (a)} The strength of causality between $y_1$ and $z_1$, measured when holding the driving signal fixed and optimizing the driven signal. Convergence occurs near iteration 10. {\bf (b)} Same as (a) but now measured when holding the driven signal fixed and optimizing the driving signal. Approximately 20 iterations were required for convergence. {\bf (c)} The strength of causality at each iteration when employing the individually optimized driving and driven signals. {\bf (d)} Same as (a) but now for the second pair of components. Convergence occurred after only 3 iterations. {\bf (e)} Rapid convergence was also found when optimizing the driving signal of the second pair. {\bf (f)}  The strength of causality in the second pair of components at each iteration. In all curves, shading represents the sem across $n=100$ realizations. }
\end{figure}

To measure the strength of causality $\mathcal{G}_{f \rightarrow g}$ between signals $f$ and $g$, the full and reduced regression models predicting $g(t)$ were explicitly learned, and the residuals then used to obtain $\mathcal{G}_{f \rightarrow g}$ via Eqn. (\ref{eqn:GCdef}). 
\paragraph{Synthetic VAR data and analysis} 
Data was generated by explicitly defining the VAR(3) system analyzed previously by Stokes and Purdon \cite{stokes2017study}:
\begin{widetext}
\begin{eqnarray}
\label{eqn:varStokes}
    \left[ \begin{array}{c}
    s_1(t) \\
    s_2(t)       \\
    s_3(t) 
    \end{array} \right] &=& \left[ \begin{array}{ccc}
        -0.9 & 0 & 0  \\
        -0.356 & 1.212 & 0 \\
        0 & -0.3098 & -1.3856
    \end{array} \right]  \left[ \begin{array}{c}
    s_1(t-1) \\
    s_2(t-1)       \\
    s_3(t-1) 
    \end{array} \right] + \nonumber \\
    && \left[ \begin{array}{ccc}
        -0.81 & 0 & 0  \\
        0.7136 & -0.49 & 0 \\
        0 & 0.50 & -0.64
    \end{array} \right]  \left[ \begin{array}{c}
    s_1(t-2) \\
    s_2(t-2)       \\
    s_3(t-2) 
    \end{array} \right] + \nonumber \\
    && \left[ \begin{array}{ccc}
        0 & 0 & 0  \\
        -0.356 & 0 & 0 \\
        0 & -0.3098 & 0
    \end{array} \right]  \left[ \begin{array}{c}
    s_1(t-3) \\
    s_2(t-3)       \\
    s_3(t-3) 
    \end{array} \right] + \left[ \begin{array}{c}
    \epsilon_1(t) \\
 \epsilon_2(t)     \\
 \epsilon_3(t)
    \end{array} \right],
\end{eqnarray}
\end{widetext}

where $\epsilon_i$, $i=1,2,3$, are independent and identically distributed innovation processes with standard deviation $\sigma=1$. $M=100$ realizations, each with a length of $N=5000$ samples, were generated by passing the vector innovation process through the impulse response (\ref{eqn:varStokes}). Projection of these latent sources to a four-dimensional observation vector followed as $\vec{x}(t)= \vec{A} \vec{s}(t)$, where the elements of 4-by-3 mixing matrix $\vec{A}$ were randomly drawn from the uniform distribution $A_{ij} \sim U[0,1]$. Notice that measurement noise enters the observed data via the innovation processes $\epsilon_i$. The proposed technique was employed to recover $P=2$ pairs of causal components:
\begin{eqnarray*}
    y_i(t) = \vec{w}_i^T \vec{x} (t), ~~~~i=1,2 \\
    z_i(t) = \vec{v}_i^T \vec{x} (t), ~~~~i=1,2,
\end{eqnarray*}
where $\vec{w}_i$ and $\vec{v}_i$ were estimated with Algorithm \ref{alg:cap}. Convergence was observed in under 20 iterations for pair 1, and under 10 for the second pair (Figure \ref{fig:convergence}).

The optimization was performed with no regularization of the block covariance matrices ($c=\infty$) and a maximum lag parameter of $L=3$. $P=2$ pairs were recovered by the optimization. When comparing the fidelity of the recovered component pairs with the ground-truth latent sources, the order of the $P=2$ pairs was corrected \emph{post hoc} if it was evident that the $(y_1,z_1)$ pair matched the $s_2 \rightarrow s_3$ relationship. In practice, the order of the recovered pairs ($s_1 \rightarrow s_2$, $s_2 \rightarrow s_3$) is insignificant, as the causal structure reflected by the two pairs is agnostic to their ordering.

To estimate the mixing matrix from the model's projection vectors $\vec{w}$ and $\vec{v}$, the driving signals $y_1$ and $y_2$, as well as the driven signal $z_2$, were individually regressed onto the observation vector $\vec{x}$. This yielded a $D$-dimensional ``forward model'' for each of the three signals, which were then compared to the three columns of the true mixing matrix. When displaying the estimated and true mixing matrix in Fig \ref{fig:sim_var}g,h, the sign and scale (L2 norm) of each estimated forward model was corrected to match that of the ground-truth mixing matrix column. 

When testing for significant differences in the strength of causality between observed signals and those recovered by the proposed method, the Wilcoxon signed rank test ($n=100$ independent VAR realizations) was employed. The maximum value across all pairs of observed signals (i.e., $\max_{i,j} \mathcal{G}_{x_i \rightarrow x_j}$) was compared against the strength of causality of the first two recovered pairs (i.e., $\mathcal{G}_{y_1 \rightarrow z_1}$, $\mathcal{G}_{y_2 \rightarrow z_2}$). The same procedure was employed to test for significant differences in the strength of causality between observed and ground-truth latent sources (i.e., $\mathcal{G}_{s_1 \rightarrow s_2}$, $\mathcal{G}_{s_2 \rightarrow s_3}$) 

\paragraph{EEG data and analysis} The neural data employed here to demonstrate the utility of proposed method has been previously described \cite{dmochowski2014audience}. Briefly, scalp EEG was collected from $n=12$ subjects freely viewing a set of 30-60 second advertisements originally broadcast during the 2012 and 2013 SuperBowl. To demonstrate the utility of the proposed method, data from a single stimulus was employed here (``Work'', Bud Light Platinum). The data was acquired with a 64-channel electrode cap connected to a BioSemi Active Two amplifier and sampled at rate of 512 Hz. A set of preprocessing steps comprised of high-pass and notch filtering, removal of eye motion artifacts by linear regression, and artifact rejection with a power criterion was applied to denoise the acquired signals. All data samples identified as artifactual by the preprocessing were linearly interpolated from neighboring samples. The interpolation allowed the computation of block covariance matrices in the presence of missing data. Moreover, data was further downsampled to a sampling frequency of 32 Hz in order to reduce the dimensionality of the ensuing block covariance matrices. The maximum lag parameter $L$ was set to 16 samples (500 ms), reflecting a tradeoff between capturing dependencies occurring on the temporal scale of neural dynamics, while avoiding excessively large covariance matrices. The number of desired component pairs was set to $P=3$.


EEG signals were mean centered prior to testing the proposed method. The block covariance matrices $\vec{\Sigma}_{1:L}$ and $\tilde{\vec{\Sigma}}$ were regularized such that the condition number of each matrix was limited to $K=10^9$. The maximum number of iterations in the grouped coordinate descent was set to 50. 

To depict the spatial topographies of the latent components, the ``forward-model'' \cite{haufe2013critical} conveying the distribution of the latent source on the scalp $\vec{a}_{w}=\vec{\Sigma}(0) \vec{w} \left( \vec{w}^T \vec{\Sigma}(0) \vec{w} \right)^{-1}$ was computed, where $\vec{\Sigma}(0)$ is the lag-zero covariance matrix of the observations $\vec{x}(t)$. Power spectra were estimated with the Thomson multitaper spectral analysis technique employing a time-bandwidth product of 64. When comparing the proposed technique with principal components analysis, the strength of causality was measured between all 90 pairs of the first 10 principal components (the approximate knee point of the data's eigenvalue spectrum). Similarly, the strength of causality was calculated among all pairs of the 10 independent components formed after performing PCA on the data. The maximum-kurtosis implementation of ICA was employed \cite{girolami1996negentropy}. 

To perform two-way ANOVA with method and component as factors, the spatial filters learned on the subject-aggregated data were applied to the recordings of individual subjects, yielding $n=12$ independent measures of the strength of causality obtained with the proposed method. The three electrode pairs with the largest (subject-aggregated) strength of causality were selected \emph{post hoc}. Similarly, the three principal and independent component pairs with the largest strength of causality were selected. The strength of causality values at the selected pairs were then measured for all subjects and employed in the ANOVA procedure. Note that the values of strength of causality yielded by the proposed method were markedly larger (i.e., $\mathcal{G}=0.32$) when evaluated on the entire (subject aggregated) data set relative to the values obtained when applying the spatial filters to individual subjects and averaging across the cohort (i.e., $\mathcal{G}=0.10 \pm 0.0031$).

\paragraph{Cryptocurrency data and analysis}
Publicly available data was obtained from an online database of historical cryptocurrency prices as captured on the Binance Exchange (\url{CryptoDataDownload.com}). Data was obtained from the following $D=19$ currencies: ADA, BAT, BNB, BTC, BTT, DASH, EOS, ETC, ETH, LINK, LTC, NEO, QTUM, TRX, USDC, XLM, XMR, XRP, and ZEC. Prices were obtained at the resolution of one minute, but subsequently downsampled by a factor of 1800 in order to capture slower dynamics manifesting across half-day segments. The opening price in each segment (i.e., as opposed to the high, low, or closing price) was employed for the analysis. 

Due to the fact that the proposed method cannot recover the scale of the latent sources, each currency's time series was standardized by removing the mean and dividing by the standard deviation. Furthermore, in order to capture genuine causal relationships unaffected by exogenous factors not captured in these currencies, the mean waveform (``global'' trend) was linearly regressed out from the multivariate time series with ordinary least squares. 

The proposed algorithm was employed with a maximum lag of $L=4$ (i.e., a two-day temporal aperture), and the $P=3$ strongest pairs of latent components were computed. Regularization of the block covariance matrices $\tilde{\vec{\Sigma}}$ and $\vec{\Sigma}_{1:L}$ was performed by limiting the condition number of both matrices to $K=1000$. 
To test for statistically significant strengths of causality in the recovered component pairs, a non-parametric test that employs surrogate data generated by randomizing the phase spectrum of the original data (while preserving its power spectrum) was employed \cite{theiler1992testing}. This procedure effectively ``shuffles'' the time series of the various cryptocurrency prices such that the genuine temporal dependencies are removed. The strength of causality measured from the surrogate records then provides a sample of the null distribution to which the true values were compared. A total of 1000 surrogate data records were formed, with the p-value measured as the number of records whose strength of causality exceeded the true value. 

To interpret the constituents of the latent souces learned by the proposed method, the elements of $\vec{w}$ and $\vec{v}$ were sorted by magnitude, and the two elements with the largest absolute value were reported in the text.

\section*{Acknowledgments}
The author would like to thank Amilcar Malave for help with figure preparation. This research was supported by the Weinbaum - Wallace H. Coulter Fund. 

\bibliographystyle{unsrt}  
\bibliography{references}  

\clearpage
\section*{Supplementary Note 1: Closed-form expressions for latent covariance} \label{note:Note1} 
In order to obtain a closed-form expression for the strength of causality $\mathcal{G}_{y \rightarrow z}$ between driving signal $y(t)$ and driven signal $z(t)$, expressions for the minimum mean squared error (MMSE) are required for both the reduced and full linear regression models:
\begin{eqnarray}
 z(t) &=& \sum_{l=1}^L h_l z(t-l) + \epsilon_r(t)    ~~~\mathrm{(reduced~model)} \\ 
    z(t) &=& \sum_{l=1}^L g_{1l} z(t-l) + \sum_{l=1}^L g_{2l} y(t-l) + \epsilon_f(t) ~~~\mathrm{(full~model)}
\end{eqnarray}
where $h_l$ are the coefficients of the temporal filter predicting the driven signal from its past in the reduced model, $g_{1l}$ are the coefficients of the filter predicting the driven signal from its own past in the full model, and $g_{2l}$ are the temporal filter weights of the filter predicting the driven signal from the past of the driving signal. The regression models can be more compactly written in vector notation as:
\begin{eqnarray}
\label{eqn:reducedVector}
    z(t) &=& \vec{h}^T \vec{z}_p + \epsilon_r(t) \\
    z(t) &=& \left[ \begin{array}{cc} \vec{g}_1 & \vec{g}_2 \end{array} \right] \left[ \begin{array}{cc} \vec{z}_p \\ \vec{y}_p \end{array} \right]   + \epsilon_f(t). \label{eqn:fullVector}
\end{eqnarray}
The coefficients of the filter $\vec{h}$ that minimizes the residual in the reduced model are given by \cite{wiener1964extrapolation}:
\begin{eqnarray}
 \label{eqn:reducedWiener}
    \vec{h} = \vec{Q}^{-1} \vec{q},
\end{eqnarray}
where $\vec{q}=E \left\{ z(t)  \vec{z}_p(t)  \right\}$ is the covariance vector between the desired signal $z$ and its own past, whose $l$th element is given by $E \left\{ z(t)  z(t-l)  \right\}$, and where $\vec{Q} = E \left\{  \vec{z}_p(t)   \vec{z}^T_p(t)  \right\}$ is the covariance matrix of $\vec{z}_p(t)$, where the element at row $i$ and column $j$ is given by $E \left\{ z(t-i)  z(t-j)  \right\} = E \left\{ z(t)  z(t-j+i)  \right\}$ under the assumption of wide-sense stationary observation data. By substituting (\ref{eqn:reducedWiener}) into (\ref{eqn:reducedVector}) and solving for the residual power, the corresponding MMSE follows as:
\begin{eqnarray}
       \Phi_r = E \{ \epsilon_r^2 \} =\sigma_z^2 - \vec{q}^T \vec{Q}^{-1}\vec{q}.
\end{eqnarray}
It is required to express $\sigma_z$, $\vec{q}$, and $\vec{Q}$ in terms of the projection vector $\vec{v}$ and the statistics of the observed data $\vec{x}(t)$. The power of the desired signal $z$ in the regression models is given by:
\begin{eqnarray}
    \sigma_z^2 = E\{ z^2(t) \} = \vec{v}^T \vec{\Sigma}(0) \vec{v},
\end{eqnarray}
where $\vec{\Sigma}(\tau)= E \left\{ \vec{x}(t) \vec{x}^T(t-\tau) \right\}$ is the lagged covariance matrix of the observed data. 
Substitution of $z(t)=\vec{v}^T \vec{x}(t)$ into the expression for $\vec{q}$ leads to:
\begin{eqnarray}
    \vec{q} &=& E \left\{ z(t) \vec{z}_p(t) \right\}  \nonumber \\
        &=&  \left( \begin{array}{c} E \left\{ z(t) z(t-1) \right\}  \\ E \left\{ z(t) z(t-2) \right\}   \\ \vdots \\ E \left\{ z(t) z(t-L)\right\}  \end{array} \right) \nonumber \\
    &=&  \left( \begin{array}{c} \vec{v}^T \vec{\Sigma}(1) \vec{v} \\ \vec{v}^T \vec{\Sigma}(2) \vec{v}  \\ \vdots \\ \vec{v}^T \vec{\Sigma}(L) \vec{v} \end{array} \right) \nonumber \\
    &=&  \left( \begin{array}{cccc} \vec{v}^T  & & & \\ & \vec{v}^T & &  \\  & & \ddots & \\ & & &  \vec{v}^T  \end{array} \right)
    \left( \begin{array}{cccc}  \vec{\Sigma}(1) &&&\\ & \vec{\Sigma}(2) & &  \\ & & \ddots & \\ & & & \vec{\Sigma}(L) \end{array} \right) 
    \left( \begin{array}{c} \vec{v} \\  \vec{v}  \\ \vdots \\  \vec{v} \end{array} \right)
    \label{eqn:qexp}.
\end{eqnarray}
By utilizing the Kronecker product $\otimes$, one can write (\ref{eqn:qexp}) as the following matrix product:
\begin{eqnarray}
       \vec{q} &=& \left( \vec{I}_L \otimes \vec{v}^T \right) \vec{\Sigma}_{1:L}  \left( \vec{1}_L \otimes \vec{v} \right),
\end{eqnarray}
where 
\begin{eqnarray*}
    \vec{\Sigma}_{1:L} = \left( \begin{array}{cccc} \vec{\Sigma}(1) & \vec{0} & \ldots  &  \vec{0} \\ 
    \vec{0} & \vec{\Sigma}(2) & \ldots & \vec{0} \\ 
    \vdots & \vec{0} & \ddots & \vdots \\ 
    \vec{0} & \ldots & \ldots & \vec{\Sigma}(L)  \\ 
    \end{array} \right) 
\end{eqnarray*}
is an $LD$-by-$LD$ block diagonal covariance matrix, $\vec{1}_L$ is a vector of all ones, and $\vec{I}_L$ is the $L$-by-$L$ identity matrix.  

Similarly, the covariance matrix $\vec{Q}$ can be written as:
\begin{eqnarray*}
    \vec{Q} &=& E \left\{ \vec{z}_p(t) \vec{z}^T_p(t)   \right\} \\
    &=&  \left[ \begin{array}{cccc} E \left\{ z^2(t-1) \right\}  &
   E \left\{ z(t-1) z(t-2) \right\} & \ldots &  E \left\{ z(t-1) z(t-L) \right\}  \\
    E \left\{ z(t-2) z(t-1) \right\} &
  E \left\{ z^2(t-2) \right\} & \ldots &   E \left\{ z(t-2) z(t-L) \right\} \\
    \vdots &
    & \ddots &
    \vdots 
     \\
     E \left\{ z(t-L) z(t-1) \right\} &
     & \ldots &   E \left\{ z^2(t-L) \right\}
 \end{array}  \right]      \nonumber \\ \\
    &=&  \left[ \begin{array}{cccc} \vec{v}^T \vec{\Sigma}(0) \vec{v} &
    \vec{v}^T \vec{\Sigma}(1) \vec{v} & \ldots & \vec{v}^T \vec{\Sigma}(L-1) \vec{v}  \\
    \vec{v}^T \vec{\Sigma}(-1) \vec{v} &
    \vec{v}^T \vec{\Sigma}(0) \vec{v} & \ldots & \vec{v}^T \vec{\Sigma}(L-2) \vec{v} ]
    \\ \vdots & & \ddots & \vdots  \\
    \vec{v}^T \vec{\Sigma}(-L+1) \vec{v} &
     & \ldots & \vec{v}^T \vec{\Sigma}(0) \vec{v} 
 \end{array}  \right]      \nonumber \\
 &=& \left( \vec{I}_L \otimes \vec{v} \right)^T \tilde{\vec{\Sigma}}  \left( \vec{I}_L \otimes \vec{v} \right),
\end{eqnarray*}
where
\begin{eqnarray*}
    \tilde{\vec{\Sigma}} = \left( \begin{array}{cccc} \vec{\Sigma}(0) & \vec{\Sigma}(-1) & \ldots  &  \vec{\Sigma}(-L+1) \\ 
    \vec{\Sigma}(1) & \vec{\Sigma}(0) & \ldots & \vec{\Sigma}(-L+2)\\ 
    \vdots & \vec{\Sigma}(1) & \ddots & \vdots \\ 
    \vec{\Sigma}(L-1) & \ldots & \ldots & \vec{\Sigma}(0)  \\ 
    \end{array} \right)
\end{eqnarray*}
is an $LD$-by-$LD$ block Toeplitz matrix. The elements of matrix $\vec{Q}$ can now be differentiated with respect to the elements of the projection vector $\vec{v}$. As described in Supplementary Note 2, this necessitates the employment of matrix differentials \cite{magnus2019matrix}. 

A similar development can be performed to derive at closed-form expressions for the covariance vector $\vec{r}$ and covariance matrix $\vec{R}$. The Wiener filter of the full regression model (\ref{eqn:reducedVector}) is given by:
\begin{eqnarray}
    \vec{g} &=& \left( \begin{array}{c} \vec{g}_1 \\  \vec{g}_2 \end{array} \right)  \nonumber \\
    &=& \vec{R}^{-1} \vec{r}, 
\end{eqnarray}
and the corresponding MMSE is given by:
\begin{eqnarray}
    \Phi_f = E \{ \epsilon_f^2 \} = \sigma_z^2 - \vec{r}^T \vec{R}^{-1}\vec{r}. 
\end{eqnarray}
Substituting $z(t)=\vec{v}^T \vec{x}(t)$ and $y(t)=\vec{w}^T\vec{x}(t)$ into the definition of $\vec{r}$ yields:
\begin{eqnarray}
    \vec{r} &=& E \left\{ z(t) \left( \begin{array}{c} \vec{z}_p(t) \\ \vec{y}_p(t) \end{array}  \right)      \right\} \nonumber \\
     &=&  \left( \begin{array}{c} E \left\{ z(t) z(t-1) \right\} \\ E \left\{ z(t) z(t-2) \right\}  \\ \vdots \\ E \left\{ z(t) z(t-L) \right\}    \\ 
E \left\{ z(t) y(t-1) \right\} \\ E \left\{ z(t) y(t-2) \right\}  \\ \vdots \\ E \left\{ z(t) y(t-L) \right\}   \end{array}  \right) \nonumber \\  
    &=&  \left( \begin{array}{c} \vec{v}^T \vec{\Sigma}(1) \vec{v} \\ \vec{v}^T \vec{\Sigma}(2) \vec{v}  \\ \vdots \\   \vec{v}^T \vec{\Sigma}(L) \vec{v}   \\
    \vec{v}^T \vec{\Sigma}(1) \vec{w} \\ \vec{v}^T \vec{\Sigma}(2) \vec{w}  \\ \vdots \\ \vec{v}^T \vec{\Sigma}(L) \vec{w}  \end{array}  \right). \label{eqn:rvecBefKron}   
\end{eqnarray}
It is straightforward to verify that (\ref{eqn:rvecBefKron}) can be factored according to:
\begin{eqnarray}
    \vec{r}= \left( \vec{I}_{2L} \otimes \vec{v}^T \right) \left( \vec{I}_2 \otimes \vec{\Sigma}_{1:L}  \right) \left( \begin{array}{c} \vec{1}_L \otimes \vec{v} \\ \vec{1}_L \otimes \vec{w} \end{array}  \right). 
\end{eqnarray}
Finally, the explicit expression for covariance matrix $\vec{R}$ is given by:
\small{
\begin{eqnarray*}
    \vec{R} &=& E \left\{ \left( \begin{array}{c} \vec{z}_p(t) \\ \vec{y}_p(t) \end{array}  \right)  \left( \begin{array}{c} \vec{z}_p(t) \\ \vec{y}_p(t) \end{array}  \right)^T      \right\}  \\
    &=&  \left( \begin{array}{cccccccc} \vec{v}^T \vec{\Sigma}(0) \vec{v} &
    \vec{v}^T \vec{\Sigma}(1) \vec{v} & \ldots & \vec{v}^T \vec{\Sigma}(L-1) \vec{v} &
    \vec{v}^T \vec{\Sigma}(0) \vec{w} &
    \vec{v}^T \vec{\Sigma}(1) \vec{w} & \ldots & \vec{v}^T \vec{\Sigma}(L-1) \vec{w}  \\
    \vec{v}^T \vec{\Sigma}(-1) \vec{v} &
    \vec{v}^T \vec{\Sigma}(0) \vec{v} & \ldots & \vec{v}^T \vec{\Sigma}(L-2) \vec{v} &
    \vec{v}^T \vec{\Sigma}(-1) \vec{w} &
    \vec{v}^T \vec{\Sigma}(0) \vec{w} & \ldots & \vec{v}^T \vec{\Sigma}(L-2) \vec{w} \\
    \vdots &
    & \ddots &
    \vdots &
    &  & \\
    \vec{v}^T \vec{\Sigma}(-L+1) \vec{v} &
     & \ldots & \vec{v}^T \vec{\Sigma}(0) \vec{v} &
    \vec{v}^T \vec{\Sigma}(-L+1) \vec{w} &
    & \ldots & \vec{v}^T \vec{\Sigma}(0) \vec{w} \\
    \vec{w}^T \vec{\Sigma}(0) \vec{v} &
    \vec{w}^T \vec{\Sigma}(1) \vec{v} & \ldots & \vec{w}^T \vec{\Sigma}(L-1) \vec{v} &
    \vec{w}^T \vec{\Sigma}(0) \vec{w} &
    \vec{w}^T \vec{\Sigma}(1) \vec{w} & \ldots & \vec{w}^T \vec{\Sigma}(L-1) \vec{w}  \\
    \vec{w}^T \vec{\Sigma}(-1) \vec{w} &
    \vec{w}^T \vec{\Sigma}(0) \vec{v} & \ldots & \vec{w}^T \vec{\Sigma}(L-2) \vec{v} &
    \vec{w}^T \vec{\Sigma}(-1) \vec{w} &
    \vec{w}^T \vec{\Sigma}(0) \vec{w} & \ldots & \vec{w}^T \vec{\Sigma}(L-2) \vec{w} \\
    \vdots &
    & \ddots &
    \vdots &
    &  & \\
    \vec{w}^T \vec{\Sigma}(-L+1) \vec{v} &
     & \ldots & \vec{w}^T \vec{\Sigma}(0) \vec{v} &
    \vec{w}^T \vec{\Sigma}(-L+1) \vec{w} &
    & \ldots & \vec{w}^T \vec{\Sigma}(0) \vec{w} 
 \end{array}  \right),      
\end{eqnarray*}
}
\normalsize
which can be factored according to:
\begin{eqnarray}
    \vec{R}=\left( \begin{array}{c} \vec{1}_2^T \otimes \vec{I}_L \otimes \vec{v}^T  \\ \vec{1}_2^T \otimes \vec{I}_L \otimes \vec{w}^T  \end{array} \right) \left( \vec{I}_2 \otimes \tilde{\vec{\Sigma}} \right) \left( \begin{array}{cc} \vec{I}_L \otimes \vec{v} & \vec{0}  \\ \vec{0} & \vec{I}_L \otimes \vec{w}  \end{array} \right).
\end{eqnarray}

\newpage
\section*{Supplementary Note 2: Gradient of objective function}
\label{sec:SI2}
The goal is to differentiate the objective function:
\begin{eqnarray}
    \mathcal{G}_{y \rightarrow z} &=& 1 - \frac{\Phi_f}{\Phi_r} \nonumber \\
    &=& 1 - \frac{\sigma_z^2 - \vec{r}^T \vec{R}^{-1}\vec{r}}{\sigma_z^2 - \vec{q}^T \vec{Q}^{-1}\vec{q}} \label{eqn:Gdef2}
\end{eqnarray}
with respect to the projection vectors $\vec{v}$ and $\vec{w}$. The derivation below relies on the chain rule, and involves the differentiation of matrices $\vec{R}$ and $\vec{Q}$ with respect to vectors $\vec{v}$ and $\vec{w}$. The reader is referred to Magnus and Neudecker \cite{magnus2019matrix} for an excellent treatment of matrix differentials, which is utilized here. 

Taking differentials of both sides of (\ref{eqn:Gdef2}) yields:
    \begin{align}
         d\mathcal{G}_{y \rightarrow z} &= \frac{\Phi_f d\Phi_r - \Phi_r d \Phi_f}{\Phi_r^2} \nonumber \\
    &= \frac{\Phi_f}{\Phi_r^2} \left( d \sigma_z^2 - d \left( \vec{q}^T \vec{Q}^{-1}\vec{q} \right) \right) - 
    \frac{1}{\Phi_r} \left( d \sigma_z^2 - d \left( \vec{r}^T \vec{R}^{-1}\vec{r} \right) \right) \nonumber \\
    &= \frac{\Phi_f}{\Phi_r^2} \left( d \sigma_z^2 - d\vec{q}^T \vec{Q}^{-1}\vec{q}  - \vec{q}^T d\vec{Q}^{-1}\vec{q} - \vec{q}^T \vec{Q}^{-1}d\vec{q} \right) 
    -
    \frac{1}{\Phi_r} \left( d \sigma_z^2 - d\vec{r}^T \vec{R}^{-1}\vec{r}  - \vec{r}^T d\vec{R}^{-1}\vec{r} - \vec{r}^T \vec{R}^{-1}d\vec{r} \right) \nonumber  \\
    &= \frac{\Phi_f}{\Phi_r^2} \left( d \sigma_z^2 - 2 \vec{q}^T \vec{Q}^{-1} d\vec{q} + \vec{q}^T \vec{Q}^{-1} d\vec{Q} \vec{Q}^{-1}\vec{q} \right) 
    -
    \frac{1}{\Phi_r} \left( d \sigma_z^2 - 2 \vec{r}^T \vec{R}^{-1} d\vec{r} + \vec{r}^T \vec{R}^{-1} d\vec{R} \vec{R}^{-1}\vec{r} \right) \nonumber \\
    &= \frac{\Phi_f}{\Phi_r^2} \left( d \sigma_z^2 - 2 \vec{q}^T \vec{Q}^{-1} d\vec{q} + \mathrm{tr}~\vec{q}^T \vec{Q}^{-1} d\vec{Q} \vec{Q}^{-1}\vec{q} \right) 
    -
    \frac{1}{\Phi_r} \left( d \sigma_z^2 - 2 \vec{r}^T \vec{R}^{-1} d\vec{r} + \mathrm{tr}~\vec{r}^T \vec{R}^{-1} d\vec{R} \vec{R}^{-1}\vec{r} \right) \nonumber \\
    &= \frac{\Phi_f}{\Phi_r^2} \left( d \sigma_z^2 - 2 \vec{q}^T \vec{Q}^{-1} d\vec{q} + \mathrm{tr}~ \vec{Q}^{-1}\vec{q} \vec{q}^T \vec{Q}^{-1} d\vec{Q} \right) 
    -
    \frac{1}{\Phi_r} \left( d \sigma_z^2 - 2 \vec{r}^T \vec{R}^{-1} d\vec{r} + \mathrm{tr}~\vec{R}^{-1}\vec{r} \vec{r}^T \vec{R}^{-1} d\vec{R}  \right) \nonumber \\
    &= \frac{\Phi_f}{\Phi_r^2} \left( d \sigma_z^2 - 2 \vec{q}^T \vec{Q}^{-1} d\vec{q} + \mathrm{vec}~ \left( \vec{Q}^{-1}\vec{q} \vec{q}^T \vec{Q}^{-1} \right)^T \mathrm{vec}~d\vec{Q} \right) 
    \nonumber \\ 
    &~~~~~~~~~~~~~~~-
    \frac{1}{\Phi_r} \left( d \sigma_z^2 - 2 \vec{r}^T \vec{R}^{-1} d\vec{r} + \mathrm{vec}~\left( \vec{R}^{-1}\vec{r} \vec{r}^T \vec{R}^{-1} \right)^T \mathrm{vec}~ d\vec{R}  \right)  \nonumber
    \\
    &= \frac{\Phi_f}{\Phi_r^2} \left( 2 \vec{v}^T \vec{\Sigma}(0) d\vec{v} - 2 \vec{q}^T \vec{Q}^{-1} \vec{J}_q d\vec{v} + \mathrm{vec} \left( \vec{Q}^{-1} \vec{q} \vec{q}^T \vec{Q}^{-1} \right)^{T} \vec{J}_Q d\vec{v} \right)  \nonumber \\ 
    &~~~~~~~~~~~~~~~- \frac{1}{\Phi_r} \left( 2 \vec{v}^T \vec{\Sigma}(0) d\vec{v} - 2 \vec{r}^T \vec{R}^{-1} \vec{J}_r \left( \begin{array}{c} d\vec{v} \\ d\vec{w} \end{array} \right) + \mathrm{vec} \left( \vec{R}^{-1} \vec{r} \vec{r}^T \vec{R}^{-1} \right)^{T} \vec{J}_R \left( \begin{array}{c} d\vec{v} \\ d\vec{w} \end{array} \right) \right), \label{eqn:dG_int} 
    \end{align}
where $\mathrm{tr}$ is the matrix trace operator, $\mathrm{vec}$ is an operator that transforms a matrix into a column vector by stacking the columns, and where the following Jacobian matrices have been defined:
\begin{eqnarray*}
d\vec{q} &=& \vec{J}_q d\vec{v} \\
\mathrm{vec}~d\vec{Q} &=& \vec{J}_Q d\vec{v} \\
d\vec{r} &=& \vec{J}_r \left( \begin{array}{c} d\vec{v} \\ d\vec{w} \end{array} \right) \\
\mathrm{vec}~d\vec{R} &=& \vec{J}_R \left( \begin{array}{c} d\vec{v} \\ d\vec{w} \end{array} \right) .
\end{eqnarray*}
Once closed-form expressions for these Jacobians are derived, it remains to substitute those expressions into (\ref{eqn:dG_int}).

\subsubsection*{Differential of $\vec{q}$}
The covariance vector $\vec{q}$ is defined by:
\begin{eqnarray}
\label{eqn:qdef2}
\vec{q}= \left( \vec{I}_L \otimes \vec{v}^T \right) \vec{\Sigma}_{1:L}  \left( \vec{1}_L \otimes \vec{v} \right).
\end{eqnarray}
The goal is to arrive at an expression of the form:
\begin{eqnarray}
\label{eqn:dvecq}
d \vec{q} = \vec{J}_{q}  d \vec{v},
\end{eqnarray}
where $\vec{J}_{q}$ is the Jacobian matrix that one seeks to identify. The following identity (often termed the ``vec'' rule) will prove useful throughout:
\begin{eqnarray*}
\mathrm{vec}(\vec{A} \vec{B} \vec{C}) = \left( \vec{C}^T \otimes  \vec{A} \right) \mathrm{vec} \left( \vec{B} \right),
\end{eqnarray*}
where matrices $\vec{A}$, $\vec{B}$, and $\vec{C}$ are defined such that the conventional matrix product $\vec{A} \vec{B} \vec{C}$ is valid. Taking differentials of both sides of (\ref{eqn:qdef2}) and applying the vec rule, one obtains:
\begin{eqnarray}
\label{eqn:dvecqint}
d\vec{q} 
&=& \left( \left( \vec{1}_L^T \otimes \vec{v}^T \right) \vec{\Sigma}_{1:L}^T \otimes \vec{I}_L \right) d ~ \mathrm{vec} \left( \vec{I}_L \otimes \vec{v}^T \right) + \left( \vec{I}_L \otimes \vec{v} \right)^T \vec{\Sigma}_{1:L} ~d~\mathrm{vec} \left( \vec{1}_L \otimes \vec{v} \right).
\end{eqnarray}
The differentials on the right-hand side may be written as \cite{magnus2019matrix}:
\begin{eqnarray}
\label{eqn:dvecILv}
d~\mathrm{vec} \left( \vec{1}_L \otimes \vec{v} \right) = \left( \vec{1}_L \otimes \vec{I}_D \right) d\vec{v},
\end{eqnarray}
and 
\begin{eqnarray}
\label{eqn:dvecILvT}
 d ~ \mathrm{vec} \left( \vec{I}_L \otimes \vec{v}^T \right) = \left( \vec{I}_L \otimes \vec{K}_{DL} \right) \left( \mathrm{vec}~\vec{I}_L \otimes \vec{I}_D \right) d \vec{v},
\end{eqnarray}
where $\vec{K}_{DL}$ is a commutation matrix satisfying:
\begin{eqnarray*}
    \vec{K}_{mn} \mathrm{vec} \left( \vec{D} \right) = \mathrm{vec} \left( \vec{D}^T \right)
\end{eqnarray*}
for $m$-by-$n$ matrix $\vec{D}$.  
In deriving (\ref{eqn:dvecILv}) and (\ref{eqn:dvecILvT}), the property on p. 206 of Magnus and Neudecker  \cite{magnus2019matrix} has been invoked to move the differential outside of the vec operator. Substituting (\ref{eqn:dvecILv}) and (\ref{eqn:dvecILvT}) into (\ref{eqn:dvecqint}), one obtains the required result:
\begin{eqnarray*}
d\vec{q}= \left[ \left( \left( \vec{1}_L^T \otimes \vec{v}^T \right) \vec{\Sigma}_{1:L}^T \otimes \vec{I}_L \right)  \left( \vec{I}_L \otimes \vec{K}_{DL} \right) \left( \mathrm{vec}~\vec{I}_L \otimes \vec{I}_D \right) 
+ \left( \vec{I}_L \otimes \vec{v} \right)^T \vec{\Sigma}_{1:L} 
\left( \vec{1}_L \otimes \vec{I}_D \right) \right] d\vec{v},
\end{eqnarray*}
where the Jacobian is identified as:
\begin{eqnarray*}
\vec{J}_{q}= \left( \left( \vec{1}_L^T \otimes \vec{v}^T \right) \vec{\Sigma}_{1:L}^T \otimes \vec{I}_L \right)  \left( \vec{I}_L \otimes \vec{K}_{DL} \right) \left( \mathrm{vec}~\vec{I}_L \otimes \vec{I}_D \right) 
+ \left( \vec{I}_L \otimes \vec{v} \right)^T \vec{\Sigma}_{1:L} 
\left( \vec{1}_L \otimes \vec{I}_D \right).
\end{eqnarray*}

\subsubsection*{Differential of $\vec{Q}$}
The $L$-by-$L$ covariance matrix of predictors in the reduced regression model is given by:
\begin{eqnarray*}
\vec{Q}= \left( \vec{I}_L \otimes \vec{v} \right)^T \tilde{\vec{\Sigma}}  \left( \vec{I}_L \otimes \vec{v} \right).
\end{eqnarray*}
The differential follows as:
\begin{eqnarray}
\label{eqn:dvecQ}
d\vec{Q}= \left( \vec{I}_L \otimes d\vec{v} \right)^T \tilde{\vec{\Sigma}}  \left( \vec{I}_L \otimes \vec{v} \right) + \left( \vec{I}_L \otimes \vec{v} \right)^T \tilde{\vec{\Sigma}}  \left( \vec{I}_L \otimes d\vec{v} \right).
\end{eqnarray}
By vectorizing both sides of (\ref{eqn:dvecQ}) and applying the vec rule to both terms on the right-hand side, one obtains:
\begin{eqnarray*}
\label{eqn:vecdQ}
\mathrm{vec}~d\vec{Q}= \left[ \left(  \tilde{\vec{\Sigma}} \left( \vec{I}_L \otimes \vec{v} \right) \right)^T \otimes \vec{I}_L \right] \mathrm{vec} \left( \vec{I}_L \otimes d\vec{v}^T \right) + \left( \vec{I}_L \otimes \left( \vec{I}_L \otimes \vec{v}^T \right) \tilde{\vec{\Sigma}} \right) \mathrm{vec} \left( \vec{I}_L \otimes d\vec{v}\right).
\end{eqnarray*}
Again using the property at the bottom of page 206 in Magnus and Neudecker \cite{magnus2019matrix}, one obtains the following expression:
\begin{eqnarray*}
\mathrm{vec}~d\vec{Q} &=& 
\left[  \left( \vec{I}_L \otimes \vec{v}^T  \right) \tilde{\vec{\Sigma}}^T  \otimes \vec{I}_L \right] \left( \vec{I}_L \otimes \vec{K}_{DL} \right) \left( \mathrm{vec}~ \vec{I}_L \otimes \vec{I}_D  \right) d\vec{v} + \\ &&
\left( \vec{I}_L \otimes \left( \vec{I}_L \otimes \vec{v}^T \right) \tilde{\vec{\Sigma}} \right) \left( \vec{I}_L \otimes \vec{I}_L \otimes \vec{I}_D \right) 
\left( \mathrm{vec}~\vec{I}_L \otimes \vec{I}_D  \right) d\vec{v},
\end{eqnarray*}
from which the Jacobian is identified as:
\begin{eqnarray*}
\vec{J}_{Q} &=& 
\left[  \left( \vec{I}_L \otimes \vec{v}^T  \right) \tilde{\vec{\Sigma}}^T  \otimes \vec{I}_L \right] \left( \vec{I}_L \otimes \vec{K}_{DL} \right) \left( \mathrm{vec}~ \vec{I}_L \otimes \vec{I}_D  \right)  + 
\left( \vec{I}_L \otimes \left( \vec{I}_L \otimes \vec{v}^T \right) \tilde{\vec{\Sigma}} \right) \left( \vec{I}_L \otimes \vec{I}_L \otimes \vec{I}_D \right) 
\left( \mathrm{vec}~\vec{I}_L \otimes \vec{I}_D  \right).
\end{eqnarray*}

\subsubsection*{Differential of r}
The $2L$-dimensional covariance vector in the full regression model is given by:
\begin{eqnarray}
\label{eqn:rvecdef2}
\vec{r}= \left( \vec{I}_{2L} \otimes \vec{v}^T \right) \left( \vec{I}_2 \otimes \vec{\Sigma}_{1:L}  \right) \left( \begin{array}{c} \vec{1}_L \otimes \vec{v} \\ \vec{1}_L \otimes \vec{w} \end{array}  \right).
\end{eqnarray}
Taking differentials of (\ref{eqn:rvecdef2}), one obtains:
\begin{eqnarray}
\label{eqn:dr2}
d\vec{r}=  \left( \vec{I}_{2L} \otimes d\vec{v}^T \right) \left( \vec{I}_2 \otimes \vec{\Sigma}_{1:L}  \right) \left( \begin{array}{c} \vec{1}_L \otimes \vec{v} \\ \vec{1}_L \otimes \vec{w} \end{array}  \right) + \left( \vec{I}_{2L} \otimes \vec{v}^T \right) \left( \vec{I}_2 \otimes \vec{\Sigma}_{1:L}  \right) \left( \begin{array}{c} \vec{1}_L \otimes d\vec{v} \\ \vec{1}_L \otimes d\vec{w} \end{array}  \right) .
\end{eqnarray}
Applying the vec operator to both sides of (\ref{eqn:dr2}) yields:
\begin{eqnarray}
\label{eqn:dr3}
\mathrm{vec}~d\vec{r}= \mathrm{vec} \left( \left( \vec{I}_{2L} \otimes d\vec{v}^T \right) \left( \vec{I}_2 \otimes \vec{\Sigma}_{1:L}  \right) \left( \begin{array}{c} \vec{1}_L \otimes \vec{v} \\ \vec{1}_L \otimes \vec{w} \end{array}  \right)  \right) + \mathrm{vec} \left( \left( \vec{I}_{2L} \otimes \vec{v}^T \right) \left( \vec{I}_2 \otimes \vec{\Sigma}_{1:L}  \right) \left( \begin{array}{c} \vec{1}_L \otimes d\vec{v} \\ \vec{1}_L \otimes d\vec{w} \end{array}  \right) \right).
\end{eqnarray}
The vec rule can now be applied to the right hand side of (\ref{eqn:dr3}):
\begin{eqnarray*}
d\vec{r}= \left( \left( \begin{array}{c} \vec{1}_L \otimes \vec{v} \\ \vec{1}_L \otimes \vec{w} \end{array}  \right)^T  \left( \vec{I}_2 \otimes \vec{\Sigma}_{1:L}  \right)^T \otimes \vec{I}_{2L} \right)
\mathrm{vec} \left( \vec{I}_{2L} \otimes \vec{dv}^T \right) +
\left( \vec{I}_{2L} \otimes \vec{v}^T \right) \left( \vec{I}_2 \otimes \vec{\Sigma}_{1:L}  \right) \left( \begin{array}{c} \mathrm{vec}~ \left( \vec{1}_L \otimes d\vec{v} \right) \\ \mathrm{vec}~ \left( \vec{1}_L \otimes d\vec{w} \right) \end{array}  \right). 
\end{eqnarray*}
Applying the property on page 206 of Magnus and Neudecker \cite{magnus2019matrix}, the resulting expression may be written as:
\begin{eqnarray*}
d\vec{r}&=& \left( \left( \begin{array}{c} \vec{1}_L \otimes \vec{v} \\ \vec{1}_L \otimes \vec{w} \end{array}  \right)^T  \left( \vec{I}_2 \otimes \vec{\Sigma}_{1:L}  \right)^T \otimes \vec{I}_{2L} \right)
\left( \vec{I}_{2L} \otimes \vec{K}_{D,2L} \right)  \left( \mathrm{vec}~\vec{I}_{2L} \otimes \vec{I}_D \right)
\vec{dv} + \nonumber \\
&&
\left( \vec{I}_{2L} \otimes \vec{v}^T \right) \left( \vec{I}_2 \otimes \vec{\Sigma}_{1:L}  \right) \left( \begin{array}{c} \left( \mathrm{vec}~ \vec{1}_L \otimes \vec{I}_D \right) d\vec{v} \\ \left( \mathrm{vec}~\vec{1}_L \otimes \vec{I}_D \right) d\vec{w} \end{array}  \right), 
\end{eqnarray*}
which can then be expressed as:
\begin{eqnarray*}
d\vec{r} &=& \left( \left( \begin{array}{c} \vec{1}_L \otimes \vec{v} \\ \vec{1}_L \otimes \vec{w} \end{array}  \right)^T  \left( \vec{I}_2 \otimes \vec{\Sigma}_{1:L}  \right)^T \otimes \vec{I}_{2L} \right)
\left( \vec{I}_{2L} \otimes \vec{K}_{D,2L} \right)  \left( \mathrm{vec}~\vec{I}_{2L} \otimes \vec{I}_D \right)
\vec{dv} + \\ &&
\left( \vec{I}_{2L} \otimes \vec{v}^T \right) \left( \vec{I}_2 \otimes \vec{\Sigma}_{1:L}  \right) 
\left( \begin{array}{cc} \vec{1}_L \otimes \vec{I}_D  & \vec{0} \\ \vec{0} & \vec{1}_L \otimes \vec{I}_D  \end{array}  \right) \left( \begin{array}{c} d\vec{v} \\ d\vec{w} \end{array}  \right).
\end{eqnarray*}
The Jacobian of $\vec{r}$ with respect to $\vec{v}$ and $\vec{w}$ can now be identified:
\begin{eqnarray*}
\vec{J}_{r} &=& \left[ \begin{array}{cc} \left( \left( \begin{array}{c} \vec{1}_L \otimes \vec{v} \\ \vec{1}_L \otimes \vec{w} \end{array}  \right)^T  \left( \vec{I}_2 \otimes \vec{\Sigma}_{1:L}  \right)^T \otimes \vec{I}_{2L} \right)
\left( \vec{I}_{2L} \otimes \vec{K}_{D,2L} \right)  \left( \mathrm{vec}~\vec{I}_{2L} \otimes \vec{I}_D \right)
& \vec{0} \end{array} \right]
+ \\ &&
\left( \vec{I}_{2L} \otimes \vec{v}^T \right) \left( \vec{I}_2 \otimes \vec{\Sigma}_{1:L}  \right) 
\left( \begin{array}{cc} \vec{1}_L \otimes \vec{I}_D  & \vec{0} \\ \vec{0} & \vec{1}_L \otimes \vec{I}_D  \end{array}  \right). 
\end{eqnarray*}

\subsubsection*{Differential of $\vec{R}$}
The covariance matrix of predictors in the full regression model is given by the $2L$-by-$2L$ matrix:
\begin{eqnarray*}
\vec{R}=\left( \begin{array}{c} \vec{1}_2^T \otimes \vec{I}_L \otimes \vec{v}^T  \\ \vec{1}_2^T \otimes \vec{I}_L \otimes \vec{w}^T  \end{array} \right) \left( \vec{I}_2 \otimes \tilde{\vec{\Sigma}} \right) \left( \begin{array}{cc} \vec{I}_L \otimes \vec{v} & \vec{0}  \\ \vec{0} & \vec{I}_L \otimes \vec{w}  \end{array} \right) .
\end{eqnarray*}
By following the same development as above, the differential of the elements of $\vec{R}$ is given by:
{
\scriptsize
\begin{align}
& \mathrm{vec}~d\vec{R} \nonumber \\
&= \left[ \left( \begin{array}{cc} \vec{I}_L \otimes \vec{v}^T & \vec{0}  \\ \vec{0} & \vec{I}_L \otimes \vec{w}^T  \end{array} \right)   \left( \vec{I}_2 \otimes \tilde{\vec{\Sigma}}^T \right) \otimes \vec{I}_{2L} \right] \left(
\vec{K}_{2,2LD} \otimes \vec{I}_L \right)^{-1} \left\{ \vec{I}_2 \otimes  \left[
\left( \vec{I}_{2L} \otimes \vec{K}_{DL} \right)  \left( \mathrm{vec}~\left(\vec{1}_2^T \otimes \vec{I}_L \right) \otimes \vec{I}_D \right) \right] \right\} \left( \begin{array}{c} d\vec{v} \\ d\vec{w} \end{array} \right)   \nonumber \\
&+ \left[ \vec{I}_{2L} \otimes \left( \begin{array}{c} \vec{1}_2^T \otimes \vec{I}_L \otimes \vec{v}^T  \\ \vec{1}_2^T \otimes \vec{I}_L \otimes \vec{w}^T \end{array} \right) \left( \vec{I}_2 \otimes \tilde{\vec{\Sigma}} \right) \right] \left( \vec{I}_2 \otimes \vec{K}_{2,L} \otimes \vec{I}_{LD} \right)^{-1}\left[ \vec{I}_{4,2} \otimes \left( \vec{I}_L \otimes \vec{I}_L \otimes \vec{I}_D \right) \left( \mathrm{vec}~\vec{I}_L \otimes \vec{I}_D \right) \right] \left( \begin{array}{c} d\vec{v} \\ d\vec{w} \end{array} \right), 
\end{align}
} 
\normalsize 
where $\vec{I}_{4,2}$ is a $4$ by $2$ matrix with ones at row 1, column 1 and at row 4, column 2, and zeros elsewhere. Thus, the Jacobian matrix $\vec{J}_{R}$ is identified as:
\begin{align}
& \vec{J}_{R}  \nonumber \\
&= \left[ \left( \begin{array}{cc} \vec{I}_L \otimes \vec{v}^T & \vec{0}  \\ \vec{0} & \vec{I}_L \otimes \vec{w}^T  \end{array} \right)   \left( \vec{I}_2 \otimes \tilde{\vec{\Sigma}}^T \right) \otimes \vec{I}_{2L} \right] \left(
\vec{K}_{2,2LD} \otimes \vec{I}_L \right)^{-1} \left\{ \vec{I}_2 \otimes  \left[
\left( \vec{I}_{2L} \otimes \vec{K}_{DL} \right)  \left( \mathrm{vec}~\left(\vec{1}_2^T \otimes \vec{I}_L \right) \otimes \vec{I}_D \right) \right] \right\}   \nonumber \\
& + \left[ \vec{I}_{2L} \otimes \left( \begin{array}{c} \vec{1}_2^T \otimes \vec{I}_L \otimes \vec{v}^T  \\ \vec{1}_2^T \otimes \vec{I}_L \otimes \vec{w}^T \end{array} \right) \left( \vec{I}_2 \otimes \tilde{\vec{\Sigma}} \right) \right] \left( \vec{I}_2 \otimes \vec{K}_{2,L} \otimes \vec{I}_{LD} \right)^{-1}\left[ \vec{I}_{4,2} \otimes \left( \vec{I}_L \otimes \vec{I}_L \otimes \vec{I}_D \right) \left( \mathrm{vec}~\vec{I}_L \otimes \vec{I}_D \right) \right]. 
\end{align}

\subsubsection*{Gradient of objective function}
Having identified the Jacobians $\vec{J}_q$, $\vec{J}_Q$, $\vec{J}_r$, and $\vec{J}_R$, the final expression for the gradient of the objective function may now assembled. { 
\begin{eqnarray}
    d\mathcal{G}_{y \rightarrow z} 
    &=& \frac{2}{\Phi_r} \left( \frac{\Phi_f}{\Phi_r} - 1 \right)  \vec{v}^T \vec{\Sigma}(0) d\vec{v} \nonumber \\
    && 
    -  \frac{2 \Phi_f}{\Phi_r^2} \vec{q}^T \vec{Q}^{-1} \vec{J}_q d\vec{v} \nonumber \\
    && + \frac{\Phi_f}{\Phi_r^2}  \mathrm{vec} \left( \vec{Q}^{-1} \vec{q} \vec{q}^T \vec{Q}^{-1} \right)^{T} \vec{J}_{Q} d\vec{v}   \nonumber \\ 
    &&  \frac{2}{\Phi_r}  \vec{r}^T \vec{R}^{-1} \vec{J}_r \left( \begin{array}{c} d\vec{v} \\ d\vec{w} \end{array} \right)  \nonumber \\ && -\frac{1}{\Phi_r}  \mathrm{vec} \left( \vec{R}^{-1} \vec{r} \vec{r}^T \vec{R}^{-1} \right)^{T} \vec{J}_R \left( \begin{array}{c} d\vec{v} \\ d\vec{w} \end{array} \right),  \label{eqn:dG_int_final}
\end{eqnarray}
}
from which one identifies the gradient of the objective function as:
\begin{align}
    & \vec{\nabla} \mathcal{G}
    = -\frac{2\mathcal{G}}{\Phi_r}  \left( \begin{array}{c}  \vec{\Sigma}(0) \vec{v} \\ \vec{0} \end{array} \right)
    -  \frac{2 \Phi_f}{\Phi_r^2} \left(  \begin{array}{c} \vec{J}_q^T \vec{Q}^{-1} \vec{q} \\ \vec{0}  \end{array} \right) + \frac{\Phi_f}{\Phi_r^2} \left( \begin{array}{c} \vec{J}_{Q}^T  \mathrm{vec} \left( \vec{Q}^{-1} \vec{q} \vec{q}^T \vec{Q}^{-1} \right) \\ \vec{0} \end{array} \right)    \nonumber \\
    & ~~~~~~~~~~~+\frac{2}{\Phi_r} \vec{J}_r^T  \vec{R}^{-1}   \vec{r}  -\frac{1}{\Phi_r} \vec{J}_R^T \mathrm{vec} \left( \vec{R}^{-1} \vec{r} \vec{r}^T \vec{R}^{-1} \right).  \label{eqn:dG_int_really_final}
    \nonumber
\end{align}

\newpage
\section*{Supplementary Note 3: Grouped coordinate descent algorithm}
To solve the optimization problem in Eqn. (\ref{eqn:GCAopt_tr}), a grouped coordinate descent algorithm was proposed. The procedure is described in Algorithm \ref{alg:cap}, where $\vec{0}$ is a vector of zeros, $\sigma$ is a small positive number, $\mathcal{N}(\vec{m},\vec{C})$ is the multivariate normal distribution with mean vector $\vec{m}$ and covariance matrix $\vec{C}$, $\vec{X}$ is a $D$-by-$T$ matrix storing the observed data, $\vec{Y}_p$ is an $L$-by-$T$ convolution matrix allowing the regression of $\vec{y}_p(t)$ onto $\vec{X}$, $\mathrm{conv matrix}$ is a routine that produces a convolution matrix, $^{\#}$ denotes the Moore-Penrose pseudoinverse, and $\vec{X}_r$ is the observed data after regressing out the contribution of the driving signal(s). The function $\mathcal{G}(\vec{w},\vec{v})$ evaluates the strength of causality (\ref{eqn:GCdef_vw}) between latent sources $y(t)=\vec{w}^T \vec{x}(t)$ and $z(t)= \vec{v}^T \vec{x}(t) $ (forward time), and $\mathcal{G}^{\mathrm{tr}}(\vec{v},\vec{w})$ evaluates the strength of causality between $y(t)=\vec{v}^T \vec{x}(-t)$ and $z(t)=\vec{w}^T \vec{x}(-t)$ (reversed time). 

\begin{algorithm}
\caption{Grouped coordinate descent for maximizing strength of causality among latent sources.}\label{alg:cap}
\begin{algorithmic}
\State $p  \gets 1 $
\While{$p \leq P$}
\State $\vec{w}_t^{(p)} \gets \mathcal{N}(\vec{0},\sigma^2 \vec{I})$
\State $\vec{v}_t^{(p)}  \gets \mathcal{N}(\vec{0},\sigma^2 \vec{I})$
    \Repeat
        \State $\vec{v}^{\ast} \gets \arg \max_{\vec{v}}$ $\mathcal{G}(\vec{w}_t^{(p)},\vec{v}) + \mathcal{G}^{\mathrm{tr}}(\vec{v},\vec{w}_t^{(p)})$
        \State $\vec{v}_t^{(p)} \gets \vec{v}^{\ast}$
        \State $\vec{w}^{\ast} \gets \arg \max_{\vec{w}} \mathcal{G}(\vec{w},\vec{v}_t^{(p)}) + \mathcal{G}^{\mathrm{tr}}(\vec{v}_t^{(p)},\vec{w})$
         \State $\vec{w}_t^{(p)} \gets \vec{w}^{\ast}$
    \Until{converged}
    \State $\vec{y}_p \gets {\vec{w}_t^{(p)}}^T \vec{X} $
    \State $\vec{Y}_p \gets \mathrm{conv matrix}\left( \vec{y}_p \right)$
    \State $\vec{X}_r \gets \vec{X} \left( \vec{I} - \vec{Y}_p^{\#} \vec{Y}_p\right) $
    \State $\vec{X} \gets \vec{X}_r$
    \State $p \gets p+1$
\EndWhile
\end{algorithmic}
\end{algorithm}
A MATLAB implementation of Algorithm \ref{alg:cap} is provided at \href{dmochow.github.io/gca}{\fontfamily{pcr}\selectfont dmochow.github.io/gca}.


\newpage
\begin{figure*}[ht!]
\centering\includegraphics[width=\linewidth]{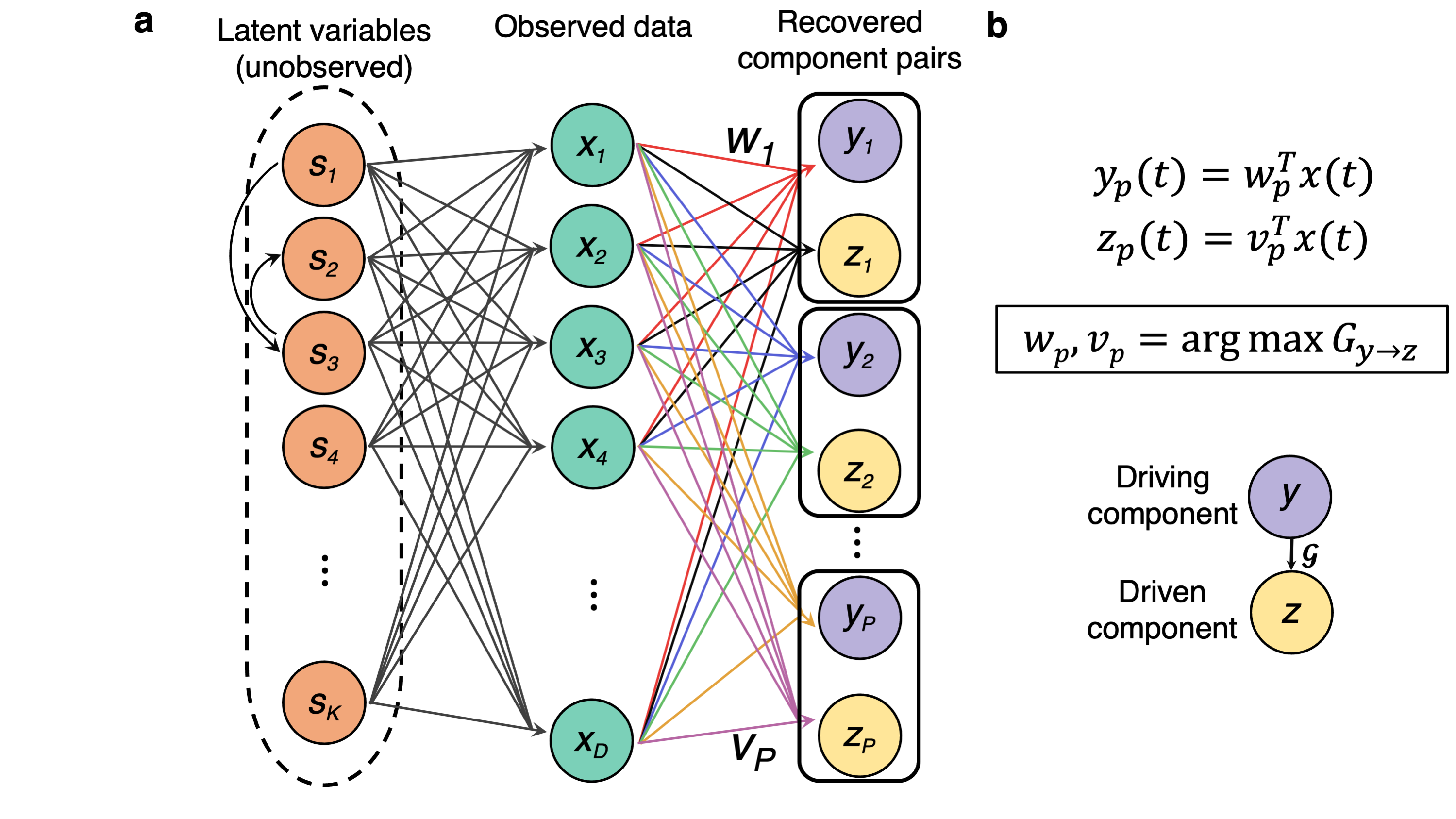}
\caption{ \label{fig:sim_var} \textbf{Extracting latent causal components from multiple time series.} {\bf(a)}  In the framework proposed here, Granger causal relations occur between pairs of latent sources whose activity governs the system dynamics. Due to a linear mixing process, the observed data manifests as a superposition of the latent sources. Consequently, direct application of Granger Causality to the observed data may not correctly identify the underlying causal structure. Instead, the approach proposed here consists of projecting the data into \emph{pairs} of components that aim to recover the directed causal connections present in the latent space. {\bf(b)}  The connection weights between the observed and recovered data are determined by solving an optimization problem that maximizes the Granger Causality between   driving signal components $y$ and driven signal components $z$. Note that although the latent components are formed instantaneously once the filters $\vec{w}$ and $\vec{v}$ are available, learning of the optimal filter weights requires observations collected over a temporal aperture of sufficient extent to estimate the data covariance. }
\end{figure*}

\begin{figure*}[ht!]
\centering\includegraphics[width=1\linewidth]{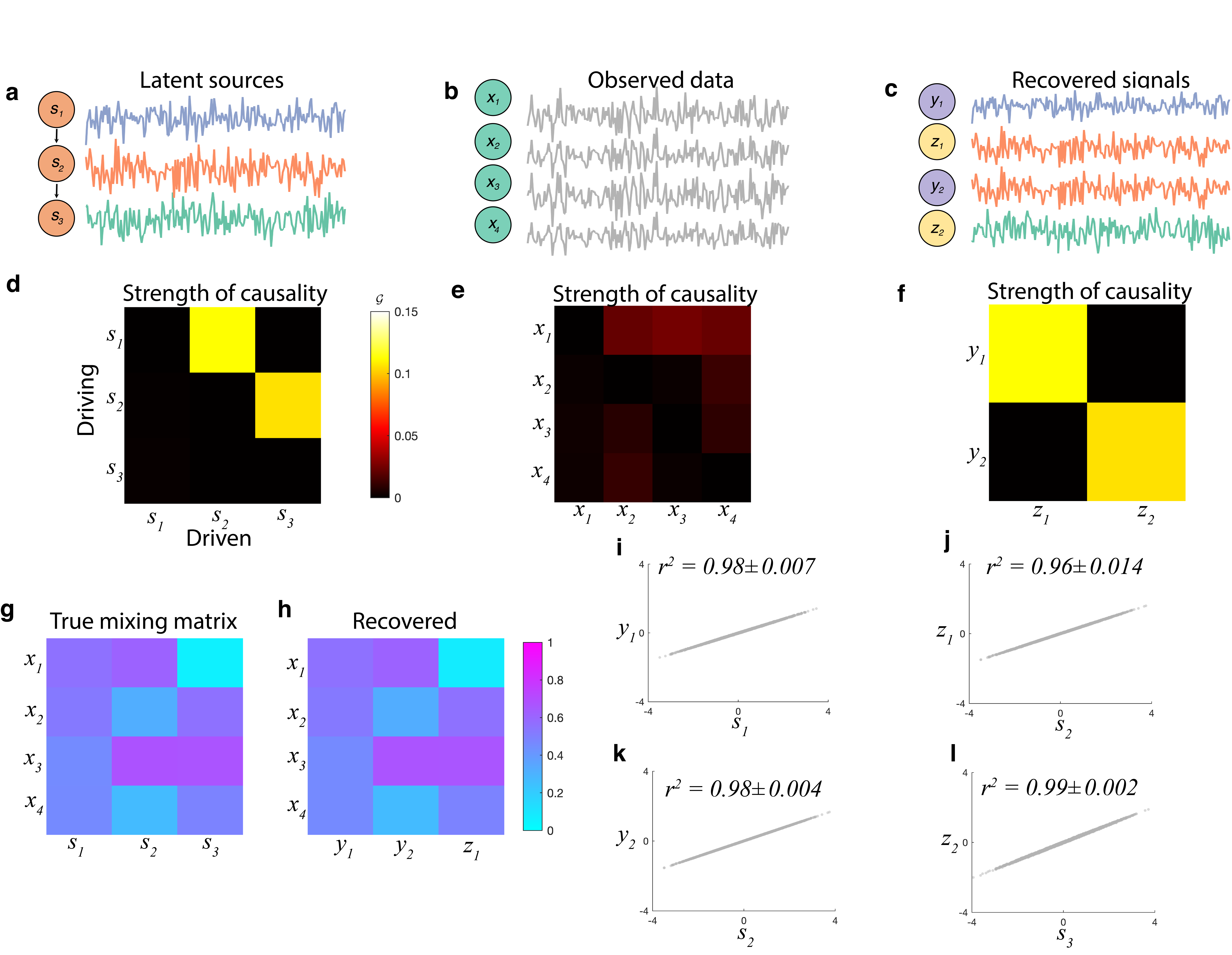}
\caption{ \label{fig:sim_var} \textbf{Identifying the latent causal structure of a three-element system.} {\bf (a)} A simple system with three latent sources was simulated, with $s_1 \rightarrow s_2$ and $s_2 \rightarrow s_3$. Data from a single realization is shown.  {\bf (b)} Projection of the source signals to a four-dimensional observation vector was modeled by a mixing matrix whose values were drawn from the uniform distribution.  {\bf (c)} Two pairs of latent sources were recovered by the proposed method.  {\bf (d)} The strength of causality between all pairs of latent sources, where it is evident that the two connections are $\mathcal{G}_{s_1 \rightarrow s_2} = 0.11 \pm 0.0011$ and $\mathcal{G}_{s_2 \rightarrow s_3} = 0.10 \pm 0.001$ (mean $\pm$ sem across 100 realizations). {\bf (e)} When computed between pairs of observed signals, the strength of causality is visibly dampened, with a maximum value of $0.066 \pm 0.002$, occurring at $x_1 \rightarrow x_3$ for the depicted realization. Notice also that the system's causal structure is not readily apparent in the observed data. {\bf (f)} The strength of causality between recovered components showed two strong connections: $\mathcal{G}_{y_1 \rightarrow z_1} = 0.11 \pm 0.0014$  ($p=4.7 \times 10^{-18}$, $n=100$, Wilcoxon signed rank test against the maximum strength of causality in the observed data) and $\mathcal{G}_{y_2 \rightarrow z_2} = 0.10 \pm 0.0010$ ($p=4.0 \times 10^{-18}$).  {\bf (g)} The mixing matrix relating the latent sources to the observed signals, shown for a single realization. {\bf (h)} The estimate of the mixing matrix as derived by regressing the recovered components onto the observed data, where $r^2=0.98 \pm 0.004$. {\bf(i)} Scatter between $s_1$ and $y_1$, where the recovered driving signal of the first pair captured $98 \pm 0.7$\% of the variability in latent source 1. {\bf(j)} Same as (i) but now shown between $s_2$ and $z_1$, where more than $96\%$ of the variability in latent source 2 was captured. {\bf(k)} The driving signal of the second pair explained over 98\% of the variability in the second latent source. Note that both $z_1$ and $y_2$ captured $s_2$. {\bf(l)} Same as (i) but now shown between $s_3$ and $z_2$, where over $99\%$ of the variability in latent source 3 was captured. }
\end{figure*}

\clearpage

\beginsupplement

\end{document}